\pdfoutput=1
\documentclass[review]{elsarticle}
\usepackage{subfigure}
\usepackage{longtable}
\usepackage{algorithm, algorithmic}
\usepackage{bm}
\usepackage{booktabs}
\usepackage{multirow}
\usepackage{geometry}
\usepackage{amsfonts,amssymb}

\usepackage{lineno,hyperref}
\modulolinenumbers[5]

\journal{arXiv}

\begin{document}

\begin{frontmatter}

\title{Self-supervised Learning for Heterogeneous Graph via Structure Information based on Metapath}


\author[mymainaddress]{Shuai Ma}
\ead{mashuai9110@163.com}

\author[mymainaddress]{Jian-wei Liu\corref{mycorrespondingauthor}}
\cortext[mycorrespondingauthor]{Corresponding author}
\ead{liujw@cup.edu.cn}

\author[mymainaddress]{Xin Zuo}
\ead{zuox@cup.edu.cn}

\address[mymainaddress]{Department of Automation, College of Information Science and Engineering,
China University of Petroleum , Beijing, Beijing, China}

\begin{abstract}
graph neural networks (GNNs) are the dominant paradigm for modeling and handling graph structure data by learning universal node representation. The traditional way of training GNNs depends on a great many labeled data, which results in high requirements on cost and time. In some special scene, it is even unavailable and impracticable. Self-supervised representation learning, which can generate labels by graph structure data itself, is a potential approach to tackle this problem. And turning to research on self-supervised learning problem for heterogeneous graphs is more challenging than dealing with homogeneous graphs, also there are fewer studies about it. In this paper, we propose a SElf-supervised learning method for heterogeneous graph via Structure Information based on Metapath (SESIM). The proposed model can construct pretext tasks by predicting jump number between nodes in each metapath to improve the representation ability of primary task. In order to predict jump number, SESIM uses data itself to generate labels, avoiding time-consuming manual labeling. Moreover, predicting jump number in each metapath can effectively utilize graph structure information, which is the essential property between nodes. Therefore, SESIM deepens the understanding of models for graph structure. At last, we train primary task and pretext tasks jointly, and use meta-learning to balance the contribution of pretext tasks for primary task. Empirical results validate the performance of SESIM method and demonstrate that this method can improve the representation ability of traditional neural networks on link prediction task and node classification task. 
\end{abstract}


\end{frontmatter}

\section{Introduction}

With the development of computer industry and the rapid growth of data volume, deep learning methods have been proposed and widely used in various fields, such as natural language processing \cite{1kim2014convolutional}, computer vision \cite{2donahue2014decaf}, and recommended system \cite{3zhu2018learning}, using end-to-end solutions by neural networks. The typical applying data structure for deep learning is Euclidean data which is characterized by regular data structure in the high dimensional feature space. However, graph as non-Euclidean data has received more and more attention in recent years, which can regard non-Euclidean structure data as nodes, and then connect the nodes among the data in the form of edges. So, graph can express the information of entities in the data and the relations between entities effectively and abstractly.

 Traditional deep learning methods cannot be applied to process graph structure data, because of the complexity of it. Graph neural networks are the hot research direction for the majority of scholars. Graph neural networks can treat graph structure data as message propagation among the connected nodes and then the dependence between nodes can be modeled, so that the graph structure data can be handled well. The essence of graph neural networks is utilizing neighbor nodes to update the feature representation of central node. So, the research work of graph neural networks started with how to fix the number of neighbor nodes and how to sort neighbor nodes, such as PATCHY-SAN \cite{4niepert2016learning}, large-scale learnable graph convolutional networks (LGCN) \cite{5gao2018large}, and diffusion-convolutional neural networks(DCNN) \cite{6atwood2016diffusion}.And now, with the further study on graph neural networks, researchers begin to turn to the research on graph neural networks in various fields of deep learning, such as graph reinforcement learning \cite{7madjiheurem2019representation}, graph transfer learning \cite{8lee2017transfer} and the explanation for graph neural networks \cite{9ying2019gnnexplainer}. However, no matter which direction graph neural networks develops, it will always be a powerful tool for handling graph structure data.

As one of the types of graphs, heterogeneous graph is a more complex graph structure, which contains different types of nodes or edges. In the real world, heterogeneous graph is a very common graph structure data, e.g., bibliographic network \cite{10hu2020strategies} and biomedical field \cite{11davis2017comparative}. Owning to the heterogeneity of nodes or edges, it is difficult for traditional graph neural networks to process heterogeneous graph. Some researchers try their best to design new neural networks for heterogeneous graph, \cite{12wang2019heterogeneous,13fu2020magnn} introduce attention mechanism to aggregate neighbor nodes in heterogeneous graph successfully and achieve best performance. \cite{14hu2020heterogeneous} proposes heterogeneous graph transformer (HGT), its whole idea is information propagation process, the metapath of which is not selected in advance. And the heterogeneous subgraph sampling algorithm is used to solve the problem that large-scale graph data cannot be applied. 

In the field of deep neural networks, we often use massive labeled data to train them to obtain better representation ability for these models. But the cost of human-annotated labels is very high, many researchers try to find alternative methods. Therefore, self-supervised learning algorithms are born at the right moment, which can utilize input data itself to generate supervised information avoiding the cost of collecting and annotating huge data. In the research about computer vision, self-supervised representation learning has achieved great success in the last years. For example, image inpainting as supervised information was proposed in \cite{15pathak2016context}, which predicts missing parts based on the rest of the image, improving the understanding for model to image through learning color and structure of it. \cite{16vondrick2018tracking} focuses on video colorization, which proposes a method for tracking large amounts of raw unlabeled data. This method is learning to establish relationship between a specified area of a color reference frame and an area of a gray frame, and then copying the color of the specified area of the reference frame to the corresponding area of the gray frame.

Since self-supervised learning has got great success in many fields, can we use it to train neural networks on graph structure data? especially for heterogeneous graph. In graph structure data, collecting and annotating data to obtain labels of nodes or edges is time-consuming and money-consuming. Although some self-supervised learning methods are studied about graph structure data by GNNs, these methods can ensure that GNNs obtain more supervised information from graph structure data itself, which has shown advantage in many downstream tasks. However, existing methods are mostly restricted to considering the common property of homogeneous graph structure data \cite{17peng2020self,18rong2020self}. Graph structure data contains a wealth of information to be exploited and mined, and they can be used as optional supervised information. Meanwhile, heterogeneous graph is the closest graph structure to real life, and the research about self-supervised learning for it is also very valuable \cite{19hwang2020self,20wang2021self}. In view of the shortcoming of existing self-supervised learning methods for GNNs, in this paper, a self-supervised method with graph structure information for heterogeneous graph based on metapath is proposed to formulate a new learning algorithm. Fig.1 is the illustration of our idea, this heterogeneous graph contains user(U), book(B) and genre(G) three types of nodes, $M$ is the metapath. U1 and U2 are probably included in several different metapaths, there may exists different jump numbers between two nodes in different metapaths. For example, 1 jump number exists in ${M_1}$,  3 jump numbers are presented in ${M_2}$ and others. This attribute can reflect the similarity and influence between two nodes and the same node containing in different metapaths reflects different semantic information. Based on this idea, we design pretext tasks predicting jump number between nodes in each metapath. SESIM can not only effectively utilize graph structure information to generate self-supervised information, but also we apply it on heterogeneous graph. Therefore, we use jump number prediction as pretext tasks to improve the representation ability of primary task, i.e., link prediction task and node classification task. The main contributions of this paper are summarized as follows:

1.We propose the SESIM model constructing supervised information from graph structure data itself to address the issue of time-consuming and money-consuming to obtain labels in GNNs. The supervised information is generated by jump number prediction in each metapath which bases on the graph structure. Therefore, it can mine the information of graph structure data itself more fully and deeper.

2.At present, there are few researches on applying self-supervised learning on heterogeneous graphs. SESIM is designed for heterogeneous graph based on metapath utilizing heterogeneous graph structure, which enables that the model can make further use of the properties of graph and leads to a more in-depth understanding of the heterogeneous graph structure.

3.We do link prediction task, node classification task and other contrast experiments using five public datasets to verity the performance of SESIM with some state-of-the-art GNNs models. The experimental results show that the pretext tasks can improve the performance of primary task and representation power of GNNs, which demonstrates the effectiveness of SESIM from various aspects.

The rest of this paper is organized as follows: Section 2 reviews traditional graph neural networks and self-supervised representation learning on general scenarios. Section 3 lists some preliminary definitions. Section 4 details the derivation process of our proposed model, depicts some key process of the self-supervised learning for heterogeneous graph, and conducts theoretical analysis. Section 5 describes the design and results for experiments, which confirms the efficiency and effectiveness of our proposed model. Section 6 concludes this paper and discuss  the future work.

\begin{figure}[!htbp]
	\centering
	\includegraphics[scale=1]{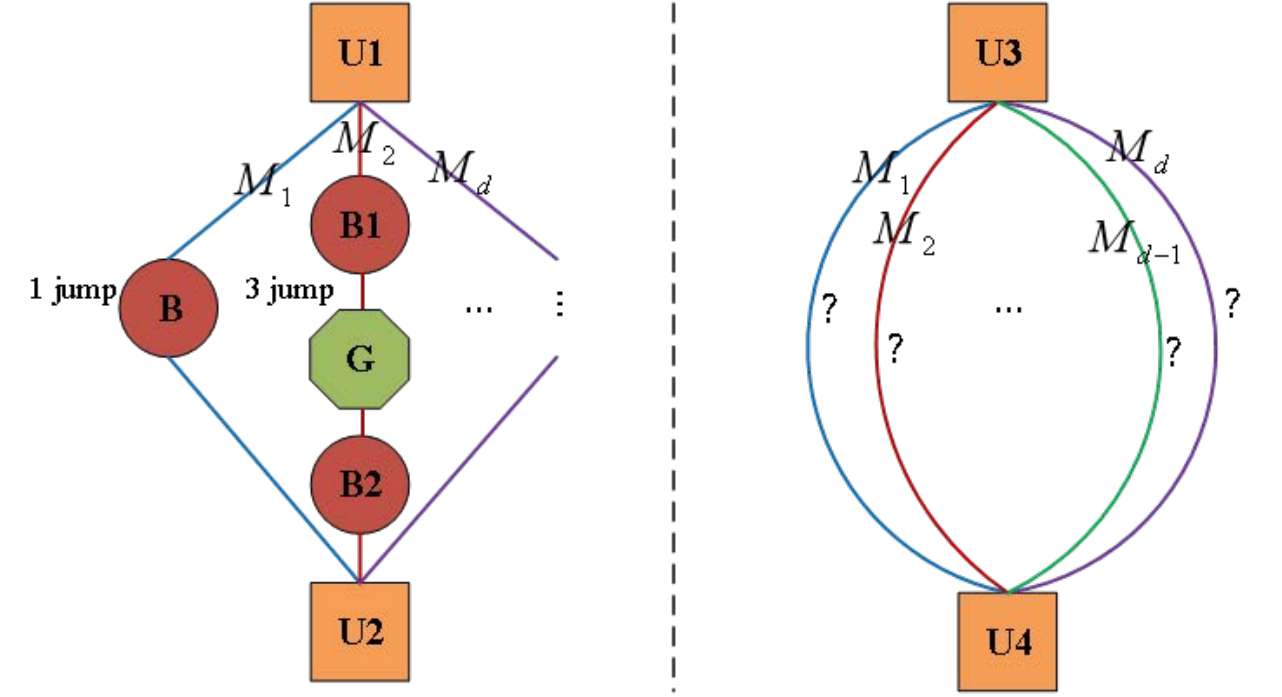}
	\caption{The illustration of our idea}
	\label{fig1}
\end{figure}

\section{Related work}

In this section, some related works are reviewed briefly, they are graph neural networks and self-supervised learning.

\subsection{Graph neural networks}

Graph structure data is becoming very common in the field of machine learning, such as nature language process, computer vision, recommended system and biochemistry, and research on how to deal with graph structure data is getting more and more important. GNNs as powerful tools to process graph structure data have become a hot research topic in recent years, which model the dependence and the message propagation between nodes in the graph.

In graph neural networks, graph convolution networks (GCNs) are the important parts, which evolve from classical convolution networks. GCNs can be classified into spectral-based algorithms and spatial-based algorithms. More specifically, spectral-based GCNs start from the spectral domain and apply the Laplacian matrix to address graph structure data. \cite{21bruna2014spectral} proposes the first generation of GCNs, which transformes classical convolu tional networks to graph cleverly by designing an effective update function for nodes. The first generation of GCNs can completely deal with the characteristics of graph structure data of unordered nodes and an uncertain number of neighbor nodes. \cite{22defferrard2016convolutional} uses K-degree Chebyshev polynomials to approximate the convolution kernel and designs a new convolution kernel, called Chebyshev network (ChebNet). Owning to ChebNet does not need to perform time-consuming eigen-decomposition operations, it can greatly improve the computational efficiency. \cite{23kipf2017semi} changes K-degree Chebyshev polynomials in \cite{22defferrard2016convolutional} into first-order representation, so that ChebNet can mine the specific graph attribute of local connection. To solve the shortcoming of the large amount of calculation of graph Fourier transform for GCNs, \cite{24xu2019graph} proposes graph wavelet neural network (GWNN) using graph wavelet transform. GWNN does not need to use the matrix decomposition with a huge amount of calculation, thus greatly improving the calculation efficiency. Spatial-based GCNs start from nodes, considering how to aggregate neighbor nodes feature efficiently. \cite{4niepert2016learning} proposes PATCHY-SAN method, containing node sequence selection, collecting a fixed-size neighbor node sets of each node and neighborhood nodes normalization three steps. GraphSAGE \cite{25hamilton2017inductive} is an inductive learning method, sampling fixed size neighbor nodes and aggregating their representations. Graph attention networks (GAT) \cite{26velivckovic2018graph} thinks different neighbor nodes with different contributions for center node, so adding attention layer in classical GNNs framework to learn the different weight. At last, using neighbor nodes with different weights to aggregate the feature representation of the central node.   

Heterogeneous graph is a complicated type of graph, some researchers also focus on it. Heterogeneous graph attention network (HAN) \cite{12wang2019heterogeneous} uses hierarchical attention mechanism to aggregate neighbor nodes feature. HAN firstly aggregates nodes feature within the metapath using attention mechanism, and then aggregates nodes feature among the metapaths by attention mechanism. Metapath aggregated graph neural network (MAGNN) \cite{13fu2020magnn} also views attention mechanism as a powerful tool to deal with heterogeneous graph, but the difference form HAN is that MAGNN considers nodes among metapaths. NSHE \cite{27zhao2020network} firstly studies the network mode that preserves heterogeneous information network embedding, which proposes a network mode sampling method to generate subgraphs and constructs multi-task learning tasks to keep the heterogeneity of each pattern instance. However, training GNNs models requires sufficient labeled data, and obtaining these labeled data is often time-consuming and money-consuming. Even, some labels require professional laboratories to conduct experiments to obtain. These models do not make full use of the data itself to generate supervised information for training GNNs models.  

\subsection{Self-supervised learning}

As is known to all, collecting and labeling massive datasets is time-consuming and costly. But the supervised information of self-supervised learning does not need to manually labeled, supervised information can be automatically constructed by pretext tasks from massive unlabeled data. And we can use this constructed supervised information to train network model, so that it can get great performance for downstream tasks.

Self-supervised learning can be classified into context-based methods, temporal-based methods and contrast-based methods. In context-based methods, \cite{28doersch2015unsupervised} explores supervised information provided by image itself. The pretext task is that selecting a pair of patches from each image after segmenting it, and then training a neural network to predict the position of one patch relative to another. The advantage of this pretext task can ensure that the feature vectors learned by primary task network has fruitful semantic information. In order to make the model really understand the meaning of the image, \cite{29pathak2016context} proposes context encoder model to generate the missing area in an image. The encoder can encode the context of an image into a latent representation and decoder predicts the missing image content using this latent representation. In temporal-based methods, (time-contrastive networks) TCN \cite{30sermanet2018time} proposes a self-supervised method based on multi-view video, whose supervised information comes from positive samples and negative samples of frames. For each frame in the video, if the feature of two frames is adjacent, they are positive. Otherwise, they are negative. The second idea starts from the perspective of multi-view, the feature of same frame coming from different views is positive, and different frames can be considered negative. \cite{31wang2015unsupervised} uses visual tracking to conduct supervised information, a track in different frames has the same feature representation, because they may belong to the same object. At last, in contrastive learning methods. Deep InfoMax (DIM) \cite{32hjelm2018learning} uses the native structure to learn representation of the image through the expression of the hidden layer. This model can have global feature and local feature at the same time and the contrastive task is classifying whether the global feature and local feature are from the same image. \cite{33he2020momentum} proposes Momentum Contrast (MoCo), which is a new dynamic dictionary mechanism for contrastive learning. There is a queue and a moving-averaged encoder in dynamic dictionary, which can maintain a large negative samples queue and do not use backpropagation to update the negative encoder. Thus, improving the expressive ability of contrastive learning.

Although self-supervised learning has got great success in a lot of fields, e.g., image and video. Nevertheless, studying self-supervised learning methods on heterogeneous graph has not yet been fully explored. So, we will investigate whether it can improve the learning capability on graph structure data in this paper.

\section{Preliminary}

\begin{figure}[!htbp]
	\centering
	\includegraphics[scale=0.8]{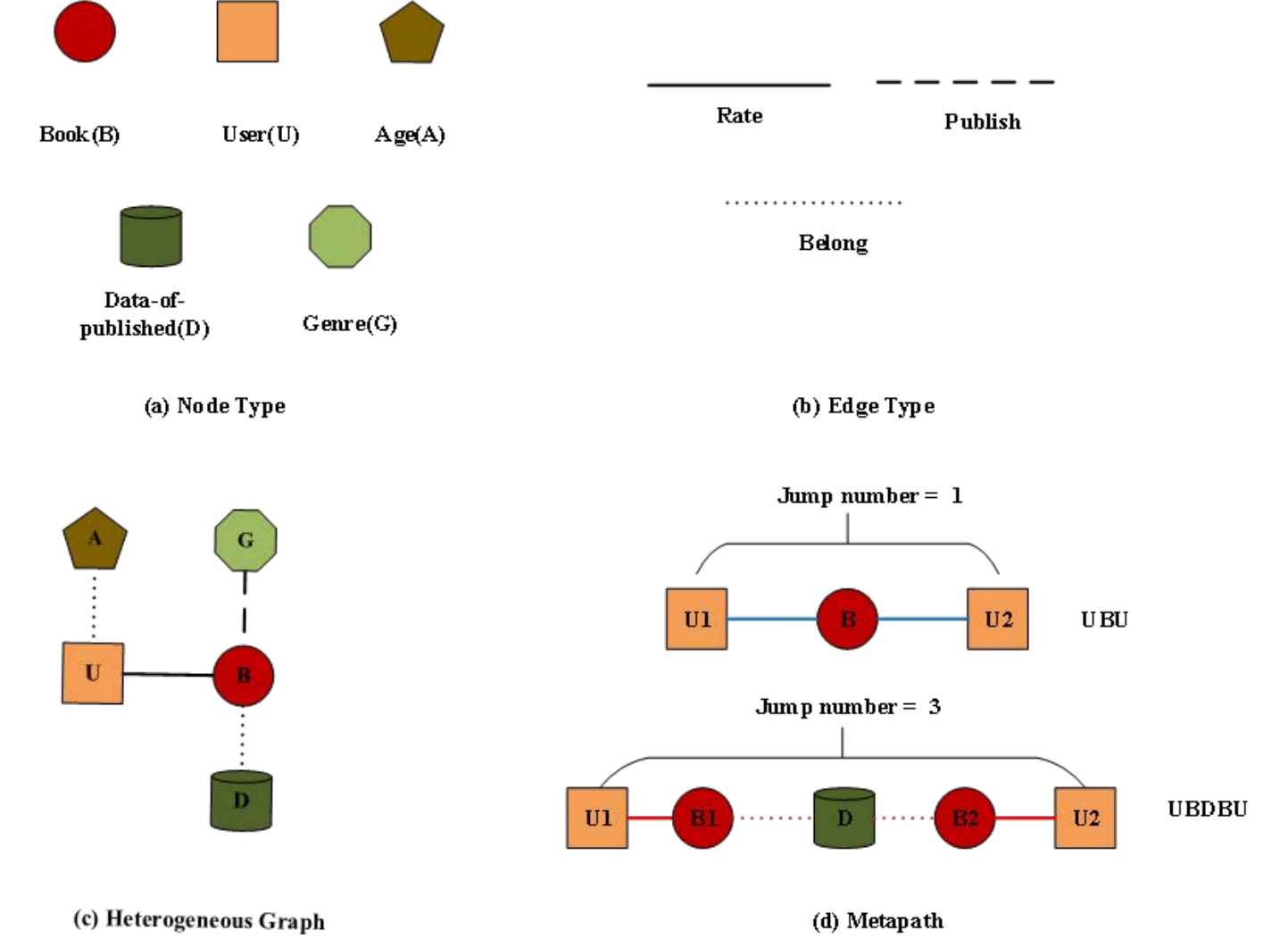}
	\caption{The preliminary of heterogeneous graph}
	\label{fig2}
\end{figure}

In this section, for readers to comprehend the proposed model clearly, we formally define three necessary concepts in connection with heterogeneous graph.

\textbf{Heterogeneous Graph.}We use $G = \left( {V,E,X,T,Q} \right)$ to represent a heterogeneous graph, $V$  is the nodes sets, $E$ is the edges sets and $X$ is the feature matrix related to $V$. $T$ and $R$ represent the types sets for $V$ and $E$. We can map node types $V$ $ \to $ $T$ and edge types $E$ $ \to $ $Q$  through two functions respectively, where $\left| T \right| + \left| Q \right| > 2$.

Fig.2 (c) is an example of heterogeneous graph. There are five types of nodes, including book (B), user (U), genre(G), data-of-published(D) and age (A). Meanwhile, there are three types of relations ("rate", "belong to" and "publish"), i.e., user rates book, and book belongs to genre.

\textbf{Metapath.}Metapath $M$ is the important attribute for heterogeneous graph G, which can be described as a path ${v_1}\mathop  \to \limits^{{R_1}} {v_2}\mathop  \to \limits^{{R_2}} ...\mathop  \to \limits^{{R_{\rm{l}}}} {v_{l + 1}}$. This path shows a composite relation $R = {R_1} \circ {R_{\rm{2}}} \circ ... \circ {{\rm{R}}_l}$ between nodes ${v_1}$ and ${v_{l + 1}}$, in which $ \circ $ is the composition operator about relation.

For example, Fig.2 (d) shows two metapaths extracted from heterogeneous graph in Fig.2 (c). UBU describes the rates of two users for the same book, and UBDBU denotes the rates of two users for two books that belong to the same genre. Multiple relations are combined in metapath, so it contains high-order structures among heterogeneous graph.

\textbf{Metapath-based Multi-jump Neighbor Nodes.}In a heterogeneous graph, assume that there are some metapaths $\left( {{M_1},{M_2},...,{M_k}} \right)$  and some nodes $\left( {{v_1},{v_2},...,{v_k}} \right)$. The metapath-based multi-jump neighbor nodes  $N_{{v_c}}^n{|_{{M_{\rm{d}}}}}$
of center node ${v_c}$ can be described as some nodes which are the neighbor nodes of center node covering $n$ jump neighbor nodes under metapath ${M_d}$.

We also take Fig.2 (d) to make a instantiation, given the metapath User-Book-User, U2 is the two-jump neighbor for U1. Similarly, the 4 jump neighbor nodes of U1 based on metapath User-Book-Data-Book-User for U2. Obviously, metapath-based multi-jump neighbor nodes can reflect different aspects of structural information in heterogeneous graph, which can be used to discern the similarity between two nodes. The metapath-based multi-jump neighbor nodes can be obtained by operating on the adjacency matrices.

\section{Proposed SESIM model}

In terms of manual labeling operation, which takes lots of time and effort or even impossible in some cases. In the era of graph structure data, especially for heterogeneous graph, devising an efficient and effective self-supervised learning method to mine the useful information from the heterogeneous graph itself is imperative for further achieving better link prediction or node classification performance. In this paper, the usual graph neural networks models and a novel self-supervised idea are combined to generate a powerful learning method SESIM for heterogeneous graph, which can take advantage of the graph structure information to improve representation ability of graph neural networks models. It can not only effectively make use of graph structure data, but also improve the performance of the primary task with the help of multiple pretext tasks. 

Graph neural networks open up new research directions for solving graph structure data, however, these methods cannot exploit supervised information from data itself to learn general node representation. Some researchers try to explore the self-supervised learning technique for graph neural networks, but in general, they mostly only utilize attribute information of graph, nevertheless the pivot graph structure information is neglected. Meanwhile, although the heterogeneous graph more commonly occurs in real-word scene, fewer self-supervised methods about it. To address these issues, the proposed algorithm SESIM considers graph structure information, i.e., jump number between nodes in each metapath for heterogeneous graph, to construct self-supervised pretext tasks. In the rest of this section, we will make a discussion about self-supervised learning method for heterogeneous graph via graph structure information based on metapath.

\subsection{Problem formulation}

\textbf{Problem definition.}Given a heterogeneous graph $G = \left( {V,E,X,T,R} \right)$, let $V$ and $\left| V \right| = n$ denote node sets and the number of nodes, respectively, $E$ and $\left| E \right| = m$ denote the edge sets and the number of edge included in the graph $G$, and $X = \left\{ {{x_1},{x_2},...,{x_n}} \right\} \in \mathbb{R}^{n \times d}$ is the corresponding feature vector sets for node sets $V = \left\{ {{v_1},{v_2},...,{v_n}} \right\}$. $M = \left\{ {{M_1},{M_2},...,{M_{\rm{d}}}} \right\}$  is metapath sets existing in heterogeneous graph $G$. The graph neural networks $h\left( {X;w} \right)$ can be utilized to project feature vector $X \in \mathbb{R}^{n \times d}$ into a low-dimensional vector space $Z = \left\{ {{z_1},{z_2},...,{z_n}} \right\} \in \mathbb{R}^{n \times l}(l < d)$ in each metapath. We should construct pseudo-labels , which are natural and intrinsic supervised information generated from data itself without any hand-annotated labels. And we use them to construct loss function $L_{pre}^{{M_d}}\left( {\hat y,f_{pre}^{{M_d}}\left( {h\left( {X;w} \right);{\theta _2}} \right)} \right)$ of pretext tasks for each metapath ${M_d}$, then we would jointly optimize the loss function of primary task and pretext tasks $\left( {{L_{pri}}\left( {y,{f_{pri}}\left( {h\left( {X;w} \right);{\theta _1}} \right)} \right) + \lambda L_{pre}^{{M_d}}\left( {\hat y,f_{pre}^{{M_d}}\left( {h\left( {X;w} \right);{\theta _2}} \right)} \right)} \right)$, where  ${f_{pri}}\left( {h\left( {X;w} \right);{\theta _1}} \right)$  is the model for primary task, $f_{pre}^{{M_d}}\left( {h\left( {X;w} \right);{\theta _2}} \right)$ is the model for pretext tasks, ${\theta _1}$ and ${\theta _2}$ are model parameters for primary task and pretext tasks. In SESIM, we aim to learn an encoder $h\left( w \right)$ of GNN, a tradeoff parameter $\lambda $, ${f_{pri}}\left( {{\theta _1}} \right)$ and $f_{pre}^{{M_d}}\left( {{\theta _2}} \right)$ for primary task and pretext tasks.

\begin{figure}[!htbp]
	\centering
	\includegraphics[scale=1]{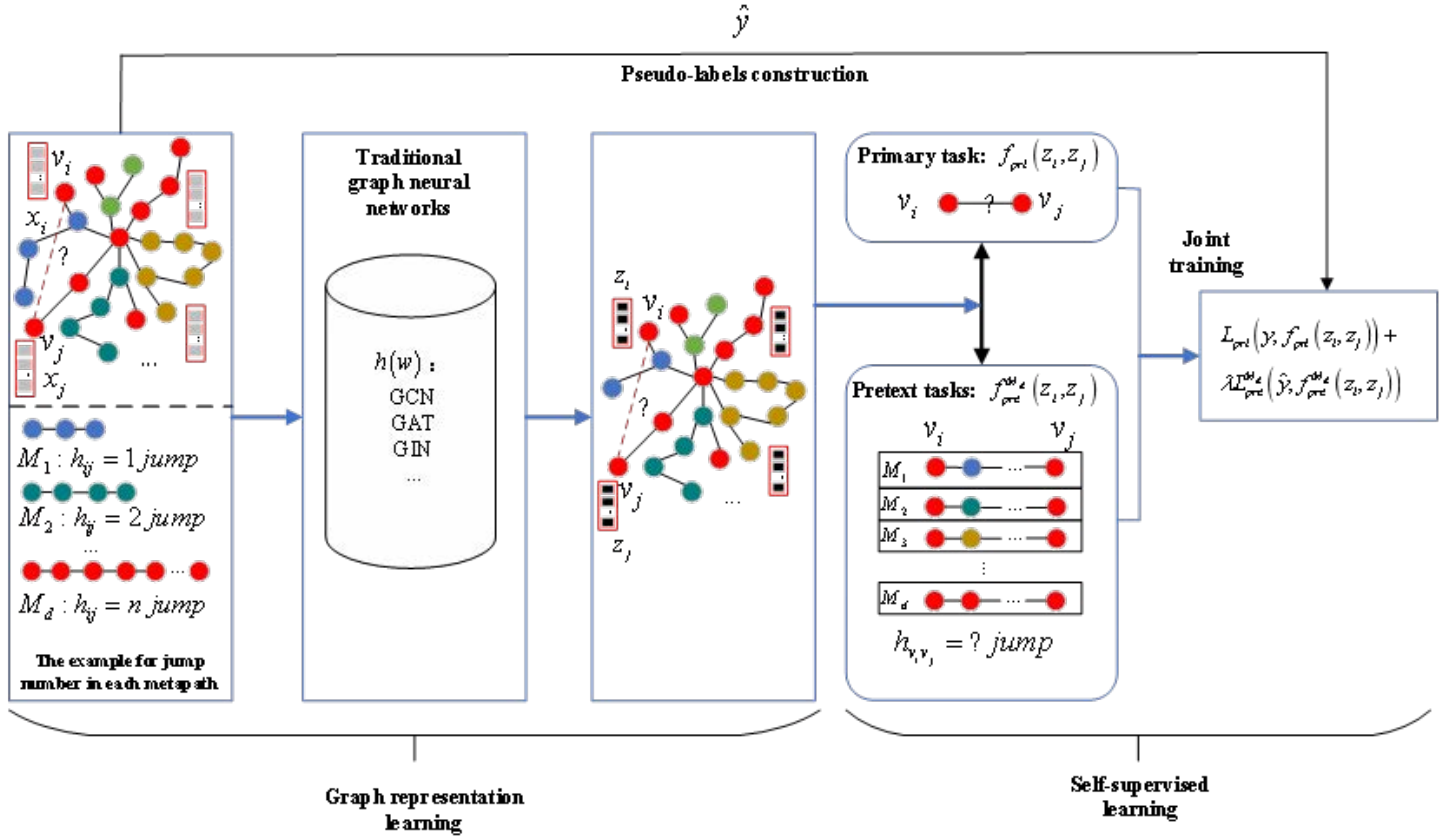}
	\caption{The architecture of SESIM}
	\label{fig3}
\end{figure}

\textbf{Framework of SESIM.}Fig.3 illustrates the flowchart of our proposed model SESIM. In this framework, the primary task is link prediction, jump number prediction for each metapath (circles with the same color represent the same metapath) are pretext tasks. Red rectangles represent the node feature vector for each node. The jump number for each metapath is shown in Fig.3. In order to clearly demonstrate the specific modeling process, we selectively pick out the node ${v_i}$ and ${v_j}$ as an example to describe the concrete building procedure of the SESIM model. In a nutshell, firstly, we should obtain self-supervised information by constructing pseudo-labels $\hat y$ for pretext tasks, the detailed construction procedures are elaborated in subsection 4.2. As illustrated in  Fig.3,  we feed node feature vectors ${x_i}$ and ${x_j}$ into graph neural networks model $h\left( {{x_i},{x_j};w} \right)$, and  get the new latent representations ${z_i}$ and ${z_j}$ for nodes ${v_i}$ and ${v_j}$, ending the graph representation learning phase. This graph neural networks $h\left( w \right)$, e.g. GCN and GAT, with the same architecture and parameters, are used to aggregate node features both for link prediction and jump number prediction. In the next self-supervised learning phase, we utilize ${z_i}$ and ${z_j}$ as input vectors of primary model ${f_{pri}}\left( {{z_i},{z_j}} \right)$  and pretext tasks $f_{pre}^{{M_d}}\left( {{z_i},{z_j}} \right)$ to get prediction values for them. We carry out joint training process to optimize simultaneously the primary task loss ${L_{pri}}\left( {y,{f_{pri}}\left( {{z_i},{z_j}} \right)} \right)$and pretext tasks loss  $L_{pre}^{{M_d}}\left( {\hat y,f_{pre}^{{M_d}}\left( {{z_i},{z_j}} \right)} \right)$ to train ${f_{pri}}\left( . \right)$, $f_{pre}^{{M_d}}\left( . \right)$ and $h\left( w \right)$, where $y$ is the labels for primary task. 

\subsection{Pseudo-labels construction}

For the self-supervised learning, constructing pseudo-labels for pretext tasks is the first step. In many cases, hand-crafted labeling process is very expensive and unavailable. We also know that learning models inherently requires massive labeled data, which led to large-scale labeled data have become one of the main barriers to the further development and deployment of learning models. During the training of pretext tasks, we also need labeled data. But these labeled data are different from traditional labeled data used in supervised learning, we call these labeled data as pseudo-label data. Pseudo-labels only come from data properties itself, that is to say pseudo-labels do not need to be labeled manually. In essence, the main difference between self-supervised learning and supervised learning is the source of labels. In supervised learning, labels come from annotators. However, in self-supervised learning, labels are obtained from data itself.

Pseudo-labels are automatically generated based on graph data properties. In the field of graph structure data, the information of nodes and edges is commonly used to construct pseudo-labels for unlabeled graph data. In SESIM, we consider the underlying graph structure information to label the data automatically, which relies on the adjacency matrix $A$. More precise, we use the jump number between nodes existing the adjacency matrix $A$ to construct pseudo-labels.

\begin{figure}[!htbp]
	\centering
	\includegraphics[scale=1]{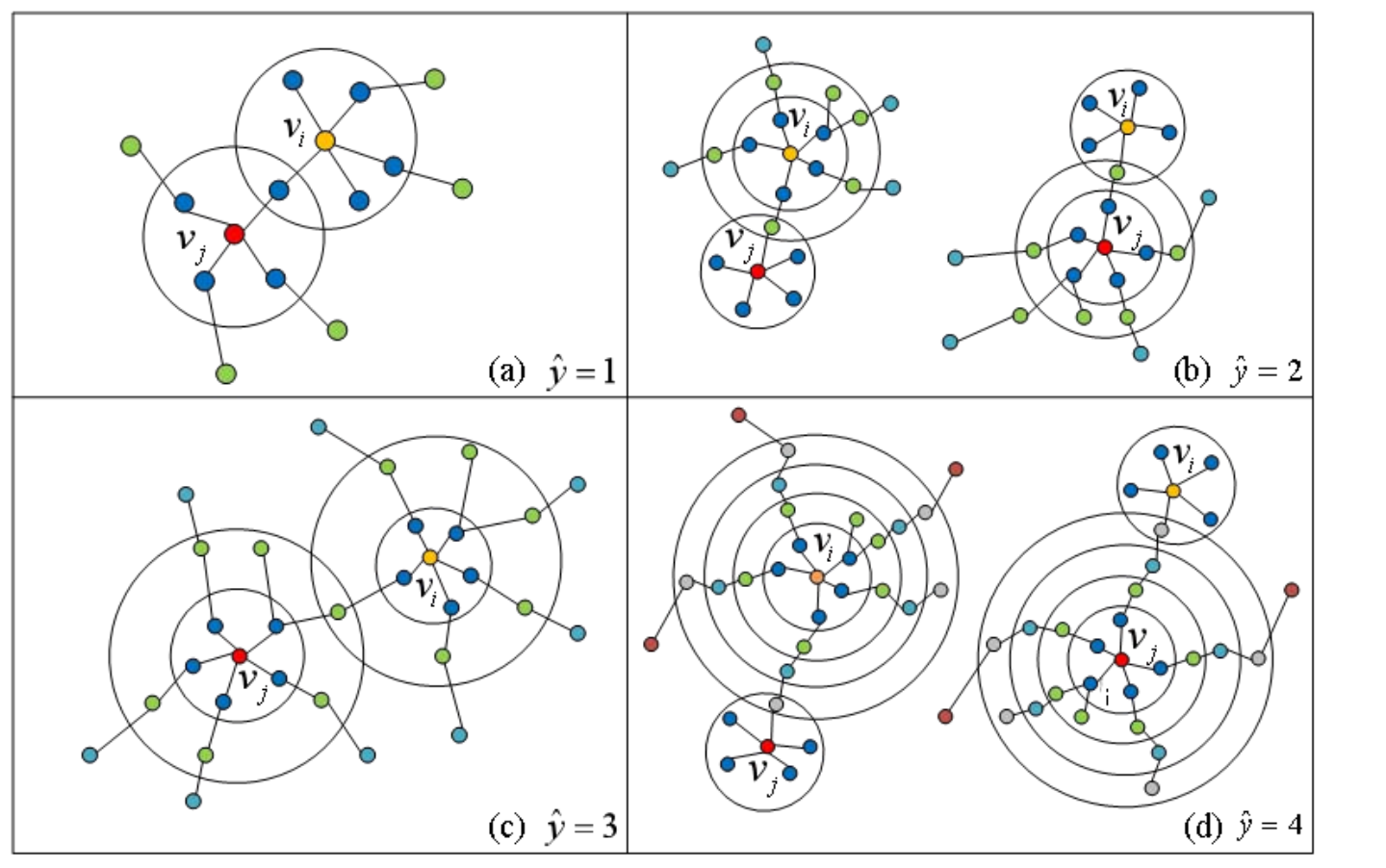}
	\caption{Illustrating the idea of constructing pseudo-labels}
	\label{fig3}
\end{figure}

To demonstrate our idea of constructing pseudo-labels from the jump number between nodes, we give an example for derived procedures of pseudo labels, i.e., the jump number, $\hat y = 1$, $\hat y = 2$, $\hat y = 3$ and $\hat y = 4$. In Fig.4, we use yellow circle and red circle to represent target nodes ${v_i}$ and ${v_j}$ in a specific metapath ${M_d}$,  respectively. Other circles with different colors represent neighbor nodes of different orders for target nodes ${v_i}$ and ${v_j}$, which are contained in different big circles. For simplicity, we only pick out the generative process of $\hat y = 1$ and $\hat y = 2$ as an example to illustrate the procedures of constructing pseudo labels, the constructing procedures for $\hat y = 3$ and $\hat y = 4$  are similar. In Fig.4.(a), we first query the adjacency matrix $A$ to obtain first-order neighbor nodes of target nodes ${v_i}$ and node ${v_j}$ in the specific metapath ${M_d}$, we use $v_i^{{N_1}}$ and $v_j^{{N_1}}$ to represent the first-order neighbor nodes sets for  nodes ${v_i}$ and node ${v_j}$, respectively. If  $v_i^{{N_1}}$ and $v_j^{{N_1}}$ include a shared first-order neighbor node for target nodes ${v_i}$ and node ${v_j}$, which is denoted with  $n_i^{\rm{1}}$ or $n_j^{\rm{1}}$, then the pseudo-label of target nodes ${v_i}$ and ${v_j}$ is set to $\hat y = 1$. Where for a neighbor node $n_l^k$, the subscript $l$ of neighbor nodes indicates the graphical node index, the superscript of  neighbor nodes indicates the order of nearest neighbors. Otherwise, we need to further figure out the information of second-order neighbor node sets of target nodes to identify the pseudo-label of target nodes, which are illustrated in Fig.4.(b). To this end, we  further query the adjacency matrix $A$ to obtain second-order neighbor nodes sets of target nodes ${v_i}$ and ${v_j}$ in the specific metapath ${M_d}$ and result in  getting the second-order neighbor nodes sets $v_i^{{N_2}}$ and $v_j^{{N_2}}$ for target nodes ${v_i}$ and ${v_j}$. Then if existing a node $n_i^2$ of $v_i^{{N_2}}$ belongs to $v_j^{{N_1}}$ or a node $n_j^2$ of $v_j^{{N_2}}$ belongs to $v_i^{{N_1}}$,  then the pseudo-label of target nodes ${v_i}$ and ${v_j}$ is set to $\hat y = 2$. The concrete formulas of constructing pseudo-labels $\hat y = 1$, $\hat y = 2$, $\hat y = 3$ and $\hat y = 4$ are as follows:

\begin{equation}
\hat y = 1{\rm{,        }}v_i^{{N_1}} \cap v_j^{{N_1}} \ne \emptyset {\rm{ }}{{\rm{|}}_{{M_d}}}{\rm{   }}
\end{equation}

\begin{equation}
\hat y = 2,{\rm{     }}n_i^2 \in {\rm{ }}v_j^{{N_1}}{\rm{ }}\parallel {\rm{  }}n_j^2 \in {\rm{ }}v_i^{{N_1}}{\rm{ }}{{\rm{|}}_{{M_d}}}
\end{equation}

\begin{equation}
\hat y = 3{\rm{,       }}v_i^{{N_2}} \cap {\rm{ }}v_j^{{N_2}} \ne \emptyset {\rm{ }}{{\rm{|}}_{{M_d}}}{\rm{   }}
\end{equation}

\begin{equation}
\hat y = 4,{\rm{     }}n_i^4 \in {\rm{ }}v_j^{{N_1}}{\rm{ }}\parallel {\rm{  }}n_j^4 \in {\rm{ }}v_i^{{N_1}}{\rm{ }}{{\rm{|}}_{{M_d}}}
\end{equation}

\begin{displaymath}
…
\end{displaymath}

where $\hat y$ is the pseudo-labels which we construct, i.e., the jump number between nodes ${v_i}$ and ${v_j}$. $v_i^{{N_1}}$, $v_i^{{N_2}}$, $v_i^{{N_4}}$, $v_j^{{N_1}}$, $v_j^{{N_2}}$ and $v_j^{{N_4}}$ are sets of the first-order neighbor nodes, the second-order neighbor nodes and the forth-order neighbor nodes for ${v_i}$ and ${v_j}$, $\parallel $ denotes the OR operation. ${\rm{ }}n_i^2$ and ${\rm{ }}n_i^4$ are second-order neighbor nodes and forth-order neighbor nodes belonging to the sets of ${\rm{ }}v_i^{{N_2}}$ and ${\rm{ }}v_i^{{N_4}}$. $n_j^2$ and $n_j^4$ are second-order neighbor node and forth-order neighbor node belonging to the sets of $v_j^{{N_2}}$ and $v_j^{{N_4}}$. ${M_d}$ is the d-th metapath.

\subsection{Jump number prediction}

Self-supervised information comes from pretext tasks, in self-supervised learning, designing effective pretext tasks is the key to improving the performance of primary task. Most existing graphical self-supervised learning approach have been focusing on common property of the graph itself. However, global graph structure is graph-specific property, and up to now there are fewer pretext tasks based on global graph structure, especially for heterogeneous graph. Meanwhile, metapath is the important tool to research heterogeneous graph, with this in mind, so for constructing effective the pretext tasks, SESIM is designed to consider global graph structure based on metapath for heterogeneous graph. These pretext tasks can let the model make better use of global graph structure, and enable the model to mine more underlying structure information, so that it will shed light on a deeper understanding of the heterogeneous graph.

\begin{figure}[!htbp]
	\centering
	\includegraphics[scale=1]{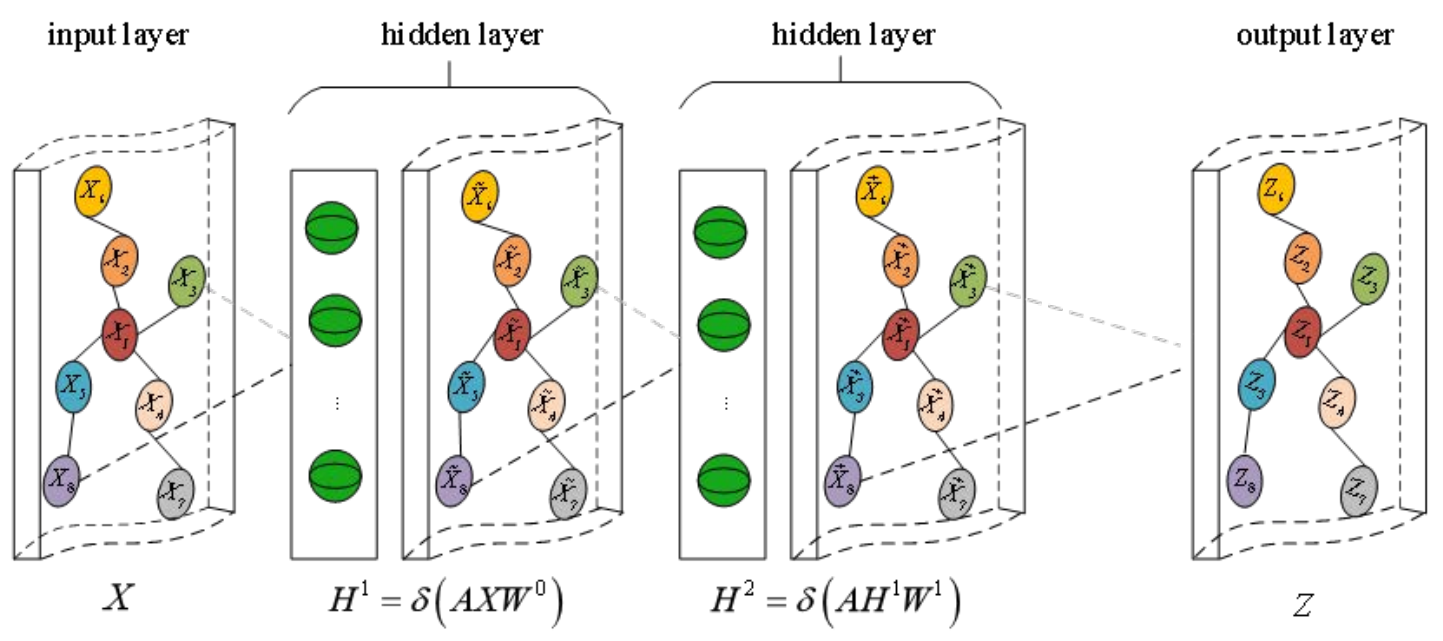}
	\caption{The architecture of GNN}
	\label{fig5}
\end{figure}

We introduce jump number prediction between ${v_i}$ and ${v_j}$ in each metapath as pretext tasks to construct self-supervised tasks. Jump number can be obtained by calculating path length of two nodes enumerating all node pairs $\left\{ {\left( {{v_i},{v_j}} \right)|{v_i},{v_j} \in V} \right\}$ under each metapath. The essential characteristics of graph-structured data is the connection relationship between nodes, we exploit adjacency matrix $A \in {\left[ {0,1} \right]^{n \times n}}$ to abstract this structural information. If there is an edge between ${v_i}$ and ${v_j}$,  ${A_{i.j}} = 1$, otherwise ${A_{i.j}} = 0$. Let $Z = \left\{ {{z_1},{z_2},...,{z_n}} \right\}$ specify the vector representation of corresponding nodes $V = \left\{ {{v_1},{v_2},...,{v_n}} \right\}$ which can be learnt by GNNs $h\left( w \right)$, such as GCN and GAT, $X = \left\{ {{x_1},{x_2},...,{x_n}} \right\}$ denotes the feature matrix for nodes $V = \left\{ {{v_1},{v_2},...,{v_n}} \right\}$. In Fig.5 and Eq. (6), we take GCN \cite{23kipf2017semi} as an example to illustrate the implementation process of GNNs. Fig.5 depicts the architecture of GCN. Eq. (6) specifies the propagation mode between layers. In Fig.5, blue spheres represent neurons. We use two-layer GCN to aggregate node features. We will introduce the detailed model parameters in subsection 5.4. The feature vector representation of each node can be computed by the following formulas:

\begin{equation}
Z = h\left( {A,X,w} \right)
\end{equation}

\begin{equation}
{H^{\left( {l + 1} \right)}} = \delta \left( {{{\hat D}^{ - \frac{1}{2}}}\hat A{{\hat D}^{ - \frac{1}{2}}}{H^{\left( l \right)}}{w^l}} \right)
\end{equation}

where $\hat D$ is the degree matrix, $\delta \left( . \right)$ is an activation function, $\hat A = A + I$ is the adjacency matrix with added self-connection, $I$ is the identity matrix, ${w^l}$ is the $l$ -th layer trainable weight matrix, ${H^{\left( l \right)}}$ is the feature latent representation in the $l$ -th hidden layer. Note that for the first hidden layer, ${H^{\left( 0 \right)}} = X$.

The prediction values of jump number can be constructed by the follow operation:

\begin{equation}
y_{{v_i},{v_j}}^{{M_d}}\left( {{\theta _2}} \right) = f_{pre}^{{M_d}}\left( {\left| {{z_i} - {z_j}} \right|;{\theta _2}} \right)
\end{equation}

where ${M_d}$ is the d-th metapath, model $f_{pre}^{{M_d}}\left( . \right)$ can predict jump number between node ${v_i}$ and node ${v_j}$ for the d-th metapath, which can linearly maps the high-dimensional vector representation to one-dimensional scalar, $|.|$ is an absolute value operation, ${\theta _2}$ is the network parameter for pretext tasks. 

Global structure information can be mined by jump number prediction based on metapath in heterogeneous graph. The method based on metapath is a divide-and-conquer approach, which divides complicated heterogeneous graph into some relatively simple subgraphs. We predict jump number in each subgraph and different jump number can reflect the “distance” between two nodes, showing their intrinsic characteristics. We make it as self-supervised information, which comes from heterogeneous graph itself, avoiding manual labeling and formulating a self-supervised representation learning algorithm for heterogeneous graph.

\subsection{Self-supervised learning}

The rationale of self-supervised learning is useing the inherent relationship of the data to label unlabeled data, thereby generating supervised information by data itself.   In former subsection 4.3, we have obtained the supervised information which comes from building pretext tasks. The training schemes of self-supervised learning are fine-tuning, unsupervised representation and joint learning \cite{34jin2020self}. Training the primary task loss and pretext tasks loss jointly is the natural idea to apply pretext tasks to improve the representation ability of primary task for heterogeneous graph. As shown in Fig.6, it illustrates the joint training scheme we used in SESIM. Motivated by \cite{19hwang2020self}, we firstly train primary task and pretext tasks jointly. Meanwhile, we use meta-learning to optimize parameters and balance the contribution of pretext tasks for primary task. At last, the learning parameters are used to derive the final objective function.

\begin{figure}[!htbp]
	\centering
	\includegraphics[scale=1]{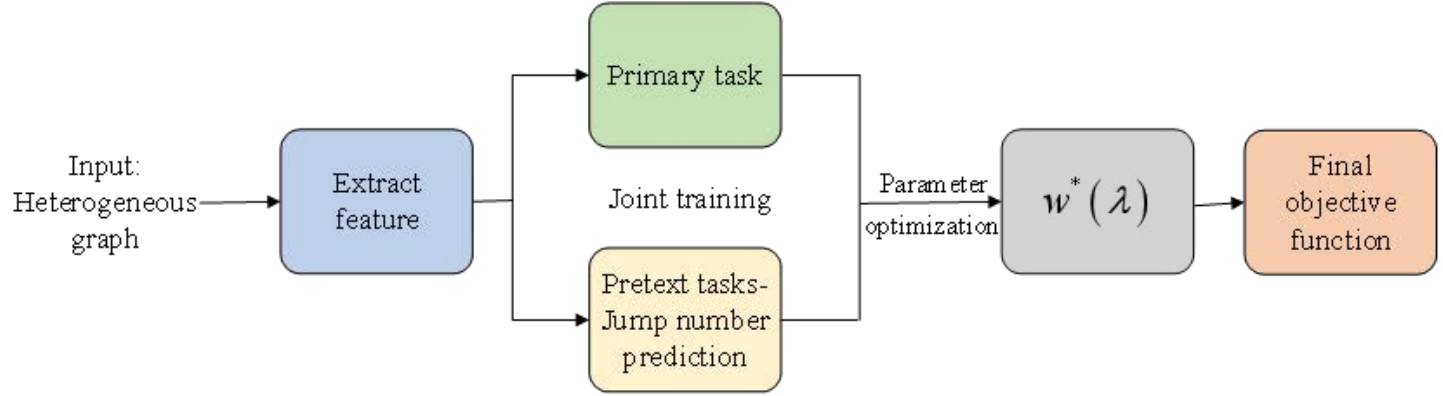}
	\caption{The joint training scheme}
	\label{fig6}
\end{figure}

In SESIM, the learning objective attempts to optimize the primary task loss function ${L_{pri}}\left( . \right)$, e.g., link prediction or node classification, and loss function $L_{pre}^d\left( . \right)$ for multiple pretext tasks, e.g., jump number prediction in each metapath, simultaneously.  And we divide the data into meta-validation data ${S^{val}}$ and training data ${S^{train}}$. The meta-validation data is used in Eq. (8) to obtain the final results. And the training data is used in Eq. (9) to optimize the intermediate parameters. This precise optimization problem can be written as:

\begin{equation}
\mathop {\min }\limits_{w,\lambda ,\theta } \sum\limits_{t = 0}^T {\frac{1}{T}{L_{pri}}\left( {y_t^{val},{f_{pri}}\left( {x_t^{val};{w^*}\left( \lambda  \right),{\theta ^*}_1} \right)} \right)} 
\end{equation}

where $T$ is the number of whole samples on meta-validation data $\left( {x_t^{val},y_t^{val}} \right)_{t = 1}^T \in {S^{val}}$, ${f_{pri}}\left( . \right)$  is the training network for primary task, e.g., link prediction. ${w^*}\left( \lambda  \right)$  is the optimal parameter which is needed to be learned from Eq.(9), ${w^*}\left( \lambda  \right)$ contains $w$ and $\lambda $ two parameters  which are the two most important parameters in SESIM. $w$ is the graph neural network model parameter for $h\left( w \right)$ to get the optimal model and $\lambda $ is the tradeoff parameter to balance primary task and pretext tasks. The detail update strategy will be described in the next subsection.

The optimal parameters ${w^*}\left( \lambda  \right)$ can be obtained as:

\begin{equation}
{w^*}\left( \lambda  \right) = \mathop {\arg \min }\limits_w \sum\limits_{i = 1}^P {\frac{1}{P}\left( \begin{array}{l}
{L_{pri}}\left( {y_i^{train},{f_{pri}}\left( {h\left( {x_i^{train};w} \right);{\theta _1}} \right)} \right) + \\
\sum\limits_{{M_d} \in M} {Con\left( {\phi _i^{{M_d}};\lambda } \right)L_{pre}^{{M_d}}\left( {\hat y_i^{train},f_{pre}^{{M_d}}\left( {h\left( {x_i^{train};w} \right);{\theta _2}} \right)} \right)} 
\end{array} \right)} 
\end{equation}

where $P$ is the number of whole samples on training data $\left( {x_i^{train},y_i^{train}} \right)_{i = 1}^P \in {S^{train}}$. ${M_d}$ is the specific metapath. $Con\left( . \right)$ is the contribution function, which can be implemented by a network. The detailed parameters setting will be shown in subsection 5.4. This function can avoid pretext tasks from playing a leading role in the learning process and ensure that pretext tasks contribute positively to primary task. $\lambda $ is the tradeoff parameter that can balance primary task and multiple pretext tasks and it is determined by meta-learning process in SESIM. $\phi _i^{{M_d}}$ is embedding representation vector for metapath ${M_d}$ of the i-th samples.$f_{pre}^{{M_d}}\left( . \right)$  is the training model in metapath ${M_d}$ for pretext tasks.

\subsection{Update parameters }

We use stochastic gradient descent (SGD) to optimize this objective function Eq. (8). During the process of optimization, we must update two parameters $w$ and $\lambda $. But it is a bi-level optimization problem, and it is difficult for us to update them directly. Motivated by \cite{35shu2019meta}, we adopted an approximate strategy to update $w$, in which SGD is used. So, we can get the updating identity for $w$ along the descent direction of primary task loss and pretext tasks loss on training data ${S^{train}}$:

\begin{equation}
{w^*}\left( \lambda  \right) \approx {\hat w^k}\left( {{\lambda ^k}} \right) = {w^k} - \alpha \frac{1}{P}\sum\limits_{i = 1}^P {\left( {\left( {{\nabla _w}L_{pri}^i\left( {{w^k},{\lambda ^k}} \right)} \right) + {\nabla _w}\left( {L_{pre}^i\left( {{w^k},{\lambda ^k}} \right)} \right)} \right)} 
\end{equation}

where $\alpha $ is the learning rate, $L_{pri}^i\left( {{w^k},{\lambda ^k}} \right) = {L_{pri}}\left( {{x_i},{y_i};{w^k},{\lambda ^k}} \right)$ and $L_{pre}^i\left( {{w^k},{\lambda ^k}} \right) = {L_{pri}}\left( {{x_i},{y_i};{w^k},{\lambda ^k}} \right)$.

$\lambda $ is an pivot parameter for self-supervised learning, when facing joint learning scheme, $\lambda $ can balance primary task and pretext tasks to enhance the representation ability for primary task. The update for $\lambda $  would help us select which pretext tasks do more contribution for primary task. From Eq. (9), we have got the update formulation for ${\hat w^k}\left( {{\lambda ^k}} \right)$.  Then parameter $\lambda $ can be updated through gradient ${\nabla _\lambda }{L_{pri}}\left( {{{\hat w}^k};{\lambda ^k}} \right)$ on a mini-batch $\left\{ {\left( {{x_t},{y_t}} \right),1 \le t \le n} \right\}$ of meta-validation data ${S^{val}}$, using the meta learning idea and iterating with the current parameter ${\lambda ^k}$:

\begin{equation}
{\lambda ^{k + 1}} = {\lambda ^k} - \beta \frac{1}{n}\sum\limits_{t = 1}^n {\left( {\left( {{\nabla _\lambda }L_{pre}^t\left( {{{\hat w}^k},{\lambda ^k}} \right)} \right)} \right)} 
\end{equation}

where $\beta $ is learning rate for $\lambda $ and $L_{pre}^t\left( {{{\hat w}^k},{\lambda ^k}} \right) = {L_{pre}}\left( {{x_t},{y_t};{{\hat w}^k},{\lambda ^k}} \right)$.

Then we use optimized parameter ${\lambda ^{k + 1}}$ received in Eq. (11) to update $w$ on a mini-batch $\left\{ {\left( {{x_i},{y_i}} \right),1 \le i \le m} \right\}$  of training data ${S^{train}}$:

\begin{equation}
{w^{k + 1}} = {w^k} - \alpha \frac{1}{m}\sum\limits_{i = 1}^m {\left( {\left( {{\nabla _w}L_{pri}^i\left( {{w^k},{\lambda ^{k + 1}}} \right)} \right) + {\nabla _w}\left( {L_{pre}^i\left( {{w^k},{\lambda ^{k + 1}}} \right)} \right)} \right)} 
\end{equation}

The precise SESIM learning procedure is shown in Algorithm 1.

\begin{algorithm}
	\renewcommand{\algorithmicrequire}{\textbf{Input:}}
	\renewcommand{\algorithmicensure}{\textbf{Output:}}
	\caption{The SESIM Learning Algorithm}
	\label{alg1}
	\begin{algorithmic}[1]
		\REQUIRE Meta-validation data ${S^{val}}$ and training data ${S^{train}}$ for primary task and pretext tasks, $P$ is the number of whole samples on training data, max iterations $K$, mini-batch size $n$ and $m$.
		\ENSURE Model parameter  ${w^k}$
		\STATE   Initialize model parameter ${w^0}$ and ${\lambda ^0}$
           \STATE \textbf{for} $k = 0$ to $K - 1$ \textbf{do}
		\STATE  $\left\{ {x_{pri}^{val},y_{pri}^{val}} \right\} \leftarrow MiniBatchSampler\left( {{S^{val}},n} \right)$
		\STATE  $\left\{ {x_{pri}^{train},y_{pri}^{train}} \right\} \leftarrow MiniBatchSampler\left( {{S^{train}},m} \right)$	
		\STATE $\left\{ {x_{pre}^{train},y_{pre}^{train}} \right\} \leftarrow MiniBatchSampler\left( {{S^{train}},m} \right)$
		\STATE Calculate  \\${w^*}\left( \lambda  \right) \approx {\hat w^k}\left( {{\lambda ^k}} \right) = {w^k} - \alpha \frac{1}{P}\sum\limits_{i = 1}^P {\left( {\left( {{\nabla _w}L_{pri}^i\left( {{w^k},{\lambda ^k}} \right)} \right) + {\nabla _w}\left( {L_{pre}^i\left( {{w^k},{\lambda ^k}} \right)} \right)} \right)} $
           \STATE   Update ${\lambda ^{k + 1}} = {\lambda ^k} - \beta \frac{1}{n}\sum\limits_{t = 1}^n {\left( {\left( {{\nabla _\lambda }L_{pre}^t\left( {{{\hat w}^k},{\lambda ^k}} \right)} \right)} \right)} $
           \STATE Update ${w^{k + 1}} = {w^k} - \alpha \frac{1}{m}\sum\limits_{i = 1}^m {\left( {\left( {{\nabla _w}L_{pri}^i\left( {{w^k},{\lambda ^{k + 1}}} \right)} \right) + {\nabla _w}\left( {L_{pre}^i\left( {{w^k},{\lambda ^{k + 1}}} \right)} \right)} \right)} $
            \STATE \textbf{end for}
	\end{algorithmic}
\end{algorithm}

\section{Experiments}

In this section, we firstly describe the evaluation metrics, the characteristic about five real-word datasets and baseline approaches used in our comparison experiments. Then we will introduce the parameter setting detailly. In the following subsections, to verify the effectiveness of our proposed method, several experiments are constructed to compare SESIM with the state-of-the-art methods on node classification task and link prediction task and discuss the experimental results. Finally, we also design some experiments to demonstrate the influence of different factors included in this self-supervised learning method.

\subsection{Evaluation metrics}

In this these experiments, to compare node classification and link prediction performance achieved by our SESIM model, we use three evaluation metrics: Macro-F1, Micro-F1 and area under curve (AUC). Before defining these evaluation metrics, confusion matrix is introduced firstly which is the basic knowledge for understanding these evaluation metrics, as shown in Table1.

\begin{table}[!htbp]
	\centering
	\caption{The confusion matrix}
	\begin{tabular}{cccc}
		\hline
		\multicolumn{2}{c}{\multirow{2}{*}{}} & \multicolumn{2}{c}{Practical results}\\ \cline{3-4}
      &      &1  &0  \\ \hline
Prediction  &1 (Positive)	 &True Positive (TP)	&False Positive (FP)\\ \cline{2-4}
	 Results &0 (Negative)	&False Negative (FN)	&True Negative (TN)\\ \hline
	\end{tabular}
\end{table}

\textbf{Precision}is the ratio of TP to predictive positive. We can use the formula TP/(TP+FP) =TP/PP to obtain it, so precision is more concerned with FP.

\textbf{Recall} is the ratio of TP to actual positive. We can use the formula TP/(TP+FN) =TP/AP to obtain it, so recall is more concerned with FN.

\textbf{F1-Score} is the harmonic mean of the precision and the recall. We can use the formula \textbf{${\rm{2}} \times {\rm{Precision}} \times {\rm{recall/}}\left( {{\rm{Precision + recall}}} \right)$} to obtain it. It is obvious that this is a combination metric of precision and Recall.

\textbf{Macro-F1:} Calculating metrics for each label and finding their unweighted mean, label imbalance should not be considered in this process. And taking the average of F1-Score as the value of Macro-F1 for all categories N.

\begin{equation}
Macro - F1 = \frac{{\sum\limits_{n = 1}^N {\left( {{F_1}1 + {F_2}1 + ...{F_n}1} \right)} }}{N}
\end{equation}

\textbf{Micro-F1:} Firstly, we calculate metrics globally by figuring up the total true positives, false negatives and false positives for all categories N. Then use them to calculate F1-Score, the concrete formulas are written as follows:

\begin{equation}
Micro - F1 = \frac{{2 \times \vec P \times \mathord{\buildrel{\lower3pt\hbox{$\scriptscriptstyle\rightharpoonup$}} 
\over R} }}{{\vec P + \mathord{\buildrel{\lower3pt\hbox{$\scriptscriptstyle\rightharpoonup$}} 
\over R} }}
\end{equation}

\begin{equation}
\vec P = \frac{{\sum\limits_{n = 1}^N {\left( {T{P_1} + T{P_2} + ... + T{P_n}} \right)} }}{{\sum\limits_{n = 1}^N {\left( {T{P_1} + T{P_2} + ... + T{P_n}} \right)}  + \sum\limits_{n = 1}^N {\left( {F{P_1} + F{P_2} + ... + F{P_n}} \right)} }}
\end{equation}

\begin{equation}
\vec R = \frac{{\sum\limits_{n = 1}^N {\left( {T{P_1} + T{P_2} + ... + T{P_n}} \right)} }}{{\sum\limits_{n = 1}^N {\left( {T{P_1} + T{P_2} + ... + T{P_n}} \right)}  + \sum\limits_{n = 1}^N {\left( {F{N_1} + F{N_2} + ... + F{N_n}} \right)} }}
\end{equation}

\textbf{AUC:} The AUC value is equal to the probability that a randomly selected positive examples is ranked higher than a randomly selected negative examples.

\subsection{Datasets}

To demonstrate the effectiveness of our proposed algorithm, we conduct the link prediction and node classification experiments respectively on public benchmark datasets which are in different areas. The detailed characteristics about these datasets are shown in Table 2 and Table 3.

\textbf{Last-FM \cite{36wang2019knowledge}:} Last-FM is a music dataset with a knowledge graph, which is released by KGNN-LS \cite{36wang2019knowledge}. This dataset contains 15084 nodes,73382 links and 122 types of edges.

\textbf{Book-Crossing \cite{37wang2018ripplenet}:} Book-Crossing is a book dataset with a knowledge graph, which is released by RippleNet \cite{37wang2018ripplenet}. This dataset contains 110739 nodes,442746 links and 52 types of edges.

\textbf{ACM \cite{12wang2019heterogeneous}:} This heterogeneous graph contains three types of nodes, which are subjects, authors and papers. The total number of nodes is 8916 and the number of each type node is {56,5835,3025}. Paper-author-paper and paper-subject-paper are metapaths in ACM. The papers are classified into three types, data mining, database and wireless communication. We define them as labels for papers. The reason for this definition depends on which conference the papers were presented at. 

\textbf{IMDB \cite{12wang2019heterogeneous}:} The heterogeneous graph structure of IMDB can be described as three types of nodes, which are directors, movies and actors, and the number of each type node is {2269,4780,5841}. Metapaths are the important attribute of heterogeneous graph, we use movie-actor-movie and movie-director-movie as metapaths in IMDB. The type of this dataset are movie and television programs, so IMDB can be classified as action, comedy, drama and we make them as labels based on their genre information. 

\textbf{DBLP \cite{12wang2019heterogeneous}:} This heterogeneous graph contains four types of nodes, which are terms, papers, conferences and authors, and the number of each type node is {8789, 14328, 20,4057}. We employ author-paper-author, author-paper-conference-paper-author and author-paper-term-paper-author as metapaths in DBLP. Four areas of database, data mining, machine learning and information retrieval are types of authors to do classification task in this experiment. This division depends on which conference does the author submits the paper to.

\begin{table}[!htbp]
	\centering
	\label{tb2}
	\caption{Datasets for link prediction}
	\begin{tabular}{cll}
		\hline
	Type	&       Last-FM             & Book-Crossing      \\ \hline
Node	&15084	&110739 \\
Edge	&73382	&442746 \\
Edge type	&122	&52 \\
 \multirow{8}{*}{Metapath} & item-user, & \\
          &user-item-user,& user-item, \\
  &item-user-item-user,&item-user-item, \\
 &user-item-artist.origin-item, &user-item-user-item, \\
 &user-item-appearing.in.film-item, &user-item-date-item-user, \\
 &user-item-instruments-item- &item-written-item-genre-item-user \\
  & artist.origin- item-user &…  \\
  &…   & \\ \hline
	\end{tabular}
\end{table}

\begin{table}[!htbp]
	\centering
	\label{tb2}
	\caption{Datasets for node prediction}
	\begin{tabular}{c|lll}
		\hline
	Type	&      ACM     & IMDB  &  DBLP    \\ \hline
\multirow{4}{*}{Node} &  &  &terms: 8789 \\
   &subjects: 56 &directors: 2269 &papers: 14328 \\
  &authors: 5835 &movies: 4780  &conferences: 20 \\
  &papers: 3025  &actors: 5841  &authors: 4057 \\
 & & &  \\
\multirow{3}{*}{Edge} & & &paper-author: 19645 \\
 &paper-author: 9744 &movie-actor: 14340 &paper-conference: 14328 \\
 &paper-subject: 3025 &movie-director: 4780 &paper-term:  88420 \\
  \multirow{5}{*}{Metapath} & & &author - paper -author \\
   &paper - author - paper &movie - actor - movie &author-paper -conference- \\
 &paper -subject- paper &movie - director - movie &paper -author \\
 & & &author-paper-term-paper- \\
 & & &author  \\ \hline
	\end{tabular}
\end{table}

\subsection{Baselines}

Considering that SESIM is a self-supervised method for heterogeneous graph based on metapath, some state-of-the-art graph neural networks models are chosen to demonstrate the effectiveness of our proposed method. They are GCN, GAT, GIN (graph isomorphism network) \cite{38xu2018powerful}, GTN (graph transformer networks) \cite{39yun2019graph}, SGC (simplifying graph convolutional networks) \cite{40wu2019simplifying}, FastGCN \cite{41chen2018fastgcn} and GWNN. More specifically, the compared methods are: 

\textbf{GCN} is a natural extension of CNN. Its essence is to extract the structural features of graph. And we can view GCN as a feature extractor, which cleverly designs a method to extract features from graph data.

\textbf{GAT} introduces attention mechanism in the procedure of aggregating nodes, it can assign different weight coefficients for each neighbor nodes according to their influence to the center node. Thus, we can get a ideal feature representation of center node.

\textbf{GIN} is a simple network architecture, which promotes the Weisfeiler-Lehman test and has the strongest discrimination ability among GNNs.

\textbf{GTN} attempts to learn the effective node representation in heterogeneous networks, the main propose of which is to identify useful connections between unconnected nodes on the original graph.

\textbf{SGC} is a simple GCN, which can be obtained by removing nonlinearities and collapsing weight matrices between consecutive layers to reduce the complexity of GCN. By analyzing of SGC model, we can get the conclusion that it is equivalent to a fixed low channel filter and a linear classifier.
 
\textbf{FastGCN} incorporates a batch training algorithm that combines importance sampling, this algorithm not only avoids dependence on test data, but also generates controllable computational consumption for each batch operation. 

\textbf{GWNN} fulfills graph wavelet transform through a fast algorithm without using a huge matrix decomposition, thus greatly improving the calculation efficiency.

\subsection{Parameter setting}

In order to reduce the influence of other factors on the experiment, the same network structure is constructed for these compared models. We design four types of jump number for pretext tasks, including $\hat y = 1$, $\hat y = 2$, $\hat y = 3$, $\hat y = 4$. And we set the number {5,5,2,2,3} for metapaths to construct subgraphs for each heterogeneous graph. Owning to the particularity of graph neural networks, we only use 2 layers neural networks with ReLu and SoftMax activation function for them. Depending on the size and attributes of the data, we set different number epochs 50 or 100 for them. The embedding dimension for all graph neural networks is set to 16 or 64. In each metapath, we first select 256 nodes as target nodes, and then sampling 8 or 16 neighbor nodes. we will utilize 40\% of the original training data for training. During the process of training, we adopt Adam optimizer \cite{42kingma2015adam} to optimize the randomly initialized parameters, the learning rate and weight-decay are set to 0.001 and 0.0001 respectively. And all of hyper-parameters are tuned on a validation set, which is set 20\% of the original training data. The detailed hyperparameter configurations for SESIM, primary task network: ${f_{pri}}\left( . \right)$ and contribution function network $Con\left( . \right)$ are described in Table 4 and Table 5.

\begin{table}[!htbp]
	\centering
	\label{tb4}
	\caption{Hyperparameter configurations for SESIM}
	\begin{tabular}{cccccc}
		\hline
 Dataset	&Last-FM	&Book-Crossing	&ACM	&IMDB &DBLP \\ \hline
Learning rate	&0.001	&0.001	&0.001	&0.001	&0.001 \\
Layers of GNN	&2	&2	&2	&2	&2 \\
Numbers of neuron	&512	&512	&512	&512	&512 \\
Weight decay	&0.0001	&0.0001	&0.0001	&0.0001	&0.0001 \\
Batch size	&256	&256	&256	&256	&256 \\
Activation function	&\multirow{2}{*}{ReLu}	&\multirow{2}{*}{ReLu}	&\multirow{2}{*}{SoftMax} &\multirow{2}{*}{SoftMax} &\multirow{2}{*}{SoftMax} \\
 in output layer & & & & & \\
\multirow{2}{*}{Loss function}	&BCE	&BCE	&CrossEntropy	&CrossEntropy	&CrossEntropy \\
  &-Loss &-Loss &-Loss &-Loss &-Loss\\ 
Numbers of epoch	&100	&50	&50	&50	&50 \\
Embedding dimension	&16	&64	&64	&64	&64 \\
Neighbor nodes size	&8	&16	&8	&8	&8 \\
Numbers of metapath	&5	&5	&2	&2	&3  \\
Training set	&40\%	&40\%	&40\%	&40\% &40\%      \\ \hline 
	\end{tabular}
\end{table}

\begin{table}[!htbp]
	\centering
	\label{tb5}
	\caption{Hyperparameter configurations for primary task network and contribution function network}
	\begin{tabular}{ccc}
		\hline
	&Primary task network:${f_{pri}}\left( . \right)$ &Contribution function \\
  & &network:$Con\left( . \right)$ \\ \hline
Learning rate	&$\alpha $=0.001	&$\beta $=0.0001 \\
Embedding dimension	&{16, 64, 64, 64, 64}	&1000 \\
Layers of network	&2	&2 \\
Numbers of neuron	&100	&1000 \\
Activation function in hidden layer	&ReLu	&ReLu \\ \hline
	\end{tabular}
\end{table}

\subsection{Link prediction results and analysis}

\begin{table}[!htbp]
	\centering
	\label{tb6}
	\caption{The link prediction results}
	\begin{tabular}{ccccccccc}
		\hline
        & \multicolumn{4}{c}{Last-FM} &\multicolumn{4}{c}{Book-Crossing} \\ \hline
   &\multicolumn{2}{c}{AUC-peak(\%)} &\multicolumn{2}{c}{AUC-mean(\%)} &\multicolumn{2}{c}{AUC-peak(\%)} &\multicolumn{2}{c}{AUC-mean(\%)} \\ \hline
	&Vanilla	&SESIM	&Vanilla	&SESIM	&Vanilla	&SESIM	&Vanilla	&SESIM \\
GCN	&78.96	&80.23	&78.56	&79.89	&70.09	&71.84	&69.76	&71.36 \\
GAT	&80.83	&81.66	&80.21	&81.39	&68.46	&69.76	&68.06	&69.16 \\
GIN	&81.66	&82.06	&81.13	&81.73	&69.56	&70.69	&69.16	&70.06 \\
GTN	&76.98	&77.93	&76.46	&77.52	&67.96	&69.35	&67.63	&68.81 \\
SGC	&75.67	&77.33	&75.06	&76.71	&68.29	&69.76	&67.83	&69.09 \\
FastGCN	&79.38	&80.68	&78.89	&80.09	&67.83	&69.35	&68.78	&68.87 \\
GWNN	&80.23	&81.94	&79.79	&81.43	&66.98	&68.34	&66.44	&67.81 \\ \hline
	\end{tabular}
\end{table}

In this part, we are going to compare the performance of SESIM and baselines on Last-FM dataset and Book-Crossing dataset for link prediction. Firstly, we carry out experiments using seven state-of-the-art graph neural networks models on these datasets to obtain the experimental results. Then, we deploy SESIM on these models to validate our idea.  AUC-peak and AUC-mean criteria are chosen to evaluate the experimental results, which is shown in Table 6. By observing it, we can get the following conclusions:

\begin{itemize}
\item Overall, both traditional graph neural networks methods and SESIM perform better on Last-FM dataset than Book-Crossing dataset. This may be determined by the nature of the dataset.
\end{itemize}

\begin{itemize}
\item The proposed model SESIM achieves the best performance in terms of AUC-peak and AUC-mean results, which is written by bold type. We can draw the conclusion that the introduction of pretext tasks can effectively improve the predictive ability of the primary task. On Last-FM dataset, our model has a large improvement on GWNN compared to other methods. But for Book-Crossing dataset, GCN which is applied SESIM gets the best.
\end{itemize}

\begin{itemize}
\item From the contrast analysis between SESIM and the other seven models, it can be seen that SESMI which uses jump number prediction based on metapath as pretext tasks, can fully mine the potential information of graph and get better link prediction performance for heterogeneous graph. Also, it demonstrates applying SESIM to traditional graph neural networks can improve their capacity of aggregating nodes and deepen their understanding of heterogeneous graph.
\end{itemize}

To illustrate the link prediction performance of these seven compared models on the validation sets of Last-FM dataset and Book-Crossing dataset more intuitively, the mean link prediction accuracies on Last-FM dataset are plotted in Fig.7, while the results of peaked link prediction accuracies and experiments on Book-Crossing dataset, which are the same as Fig.7, are no more plotted. In Fig.7, the abscissa represents the proportion of training dataset to whole data, and the ordinate represents the mean link prediction accuracies (AUC-mean).

Fig.7 visually shows the link prediction accuracies of seven compared models on the validation sets of the Last-FM dataset. From Fig.7, we can observe some phenomena that overall as the increase of the proportion of training dataset to whole data, all results are on the rise. In detail, from 20 to 60 percent, the prediction accuracies increase faster, while from 60 to 80 percent, the growth rates for the prediction accuracies increase slower. It demonstrates that a great many of training data can make the model have better learning ability. Furthermore, the AUC-mean curve for the proposed SESIM, which is marked with a solid red line, is above the other curves in Fig.7(a), Fig.7(b), Fig.7(c) and Fig.7(d). This means that the pretext tasks we designed can effectively enhance the performance of primary task.  And it is shown although traditional state-of-the-art graph neural networks have strong processing ability to heterogeneous graph data, reasonable self-supervised information can improve its aggregation operation of nodes. Also, the pretext tasks for jump number prediction can sufficiently mine the potential information hiding in the graph structure.

\begin{figure}[htbp]
\centering
\subfigure[]
{
    \begin{minipage}[b]{.8\linewidth}
        \centering
        \includegraphics[scale=0.45]{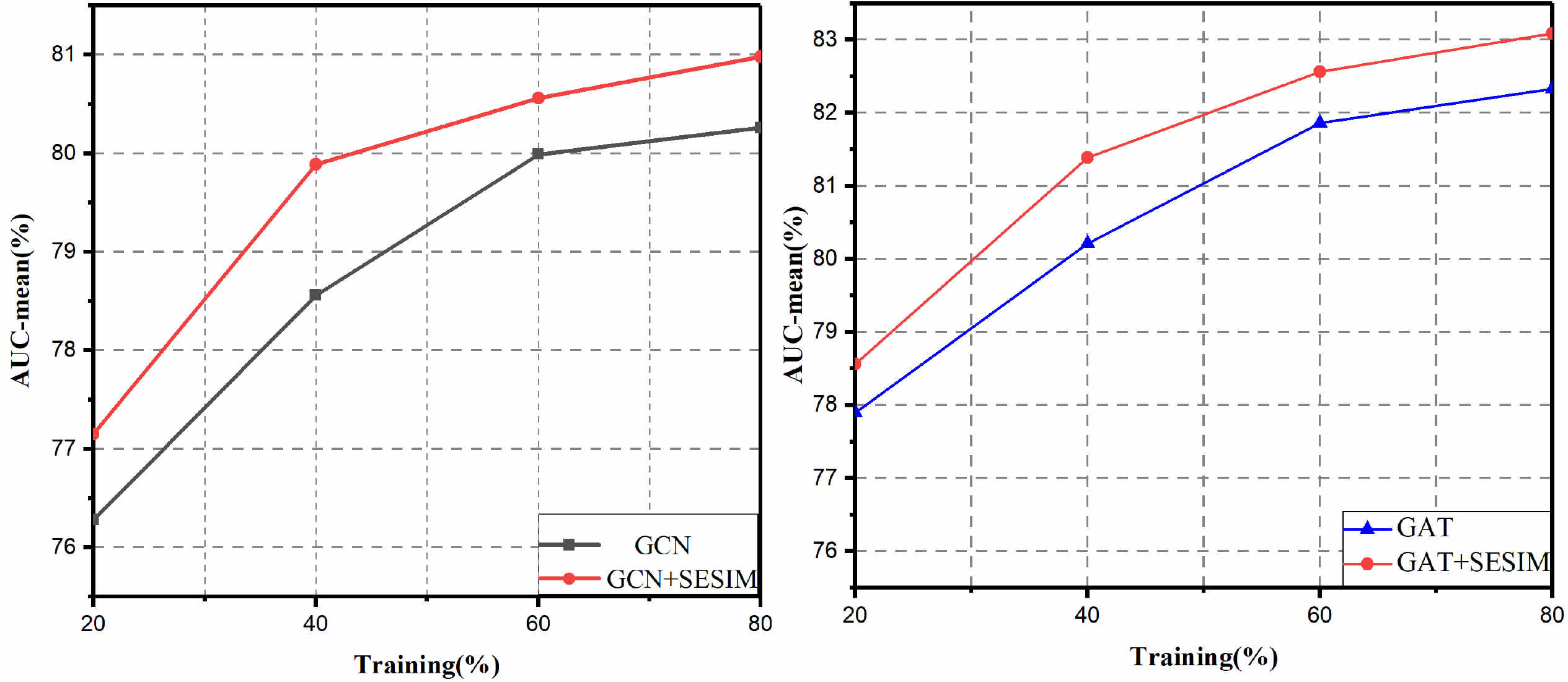}
    \end{minipage}
}
\subfigure[]
{
 	\begin{minipage}[b]{.8\linewidth}
        \centering
        \includegraphics[scale=0.51]{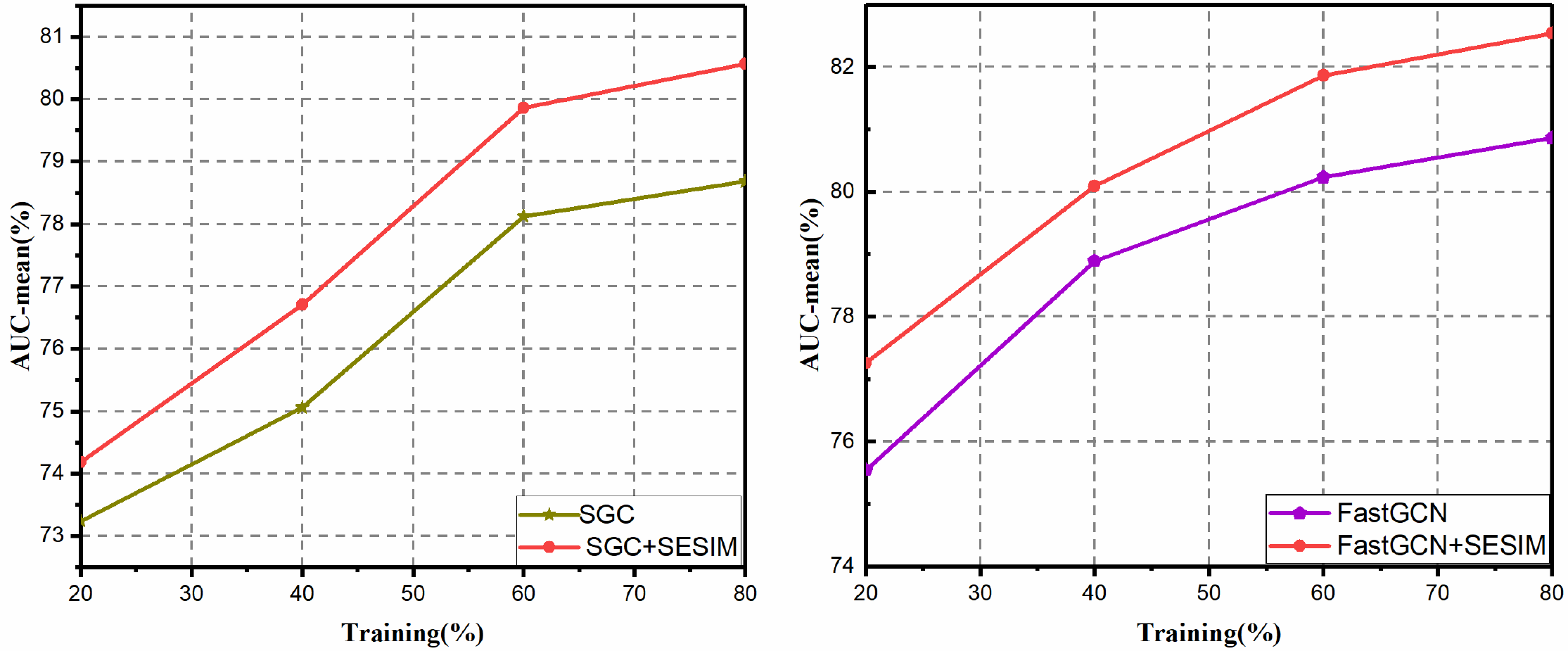}
    \end{minipage}
}
\subfigure[]
{
 	\begin{minipage}[b]{.8\linewidth}
        \centering
        \includegraphics[scale=0.57]{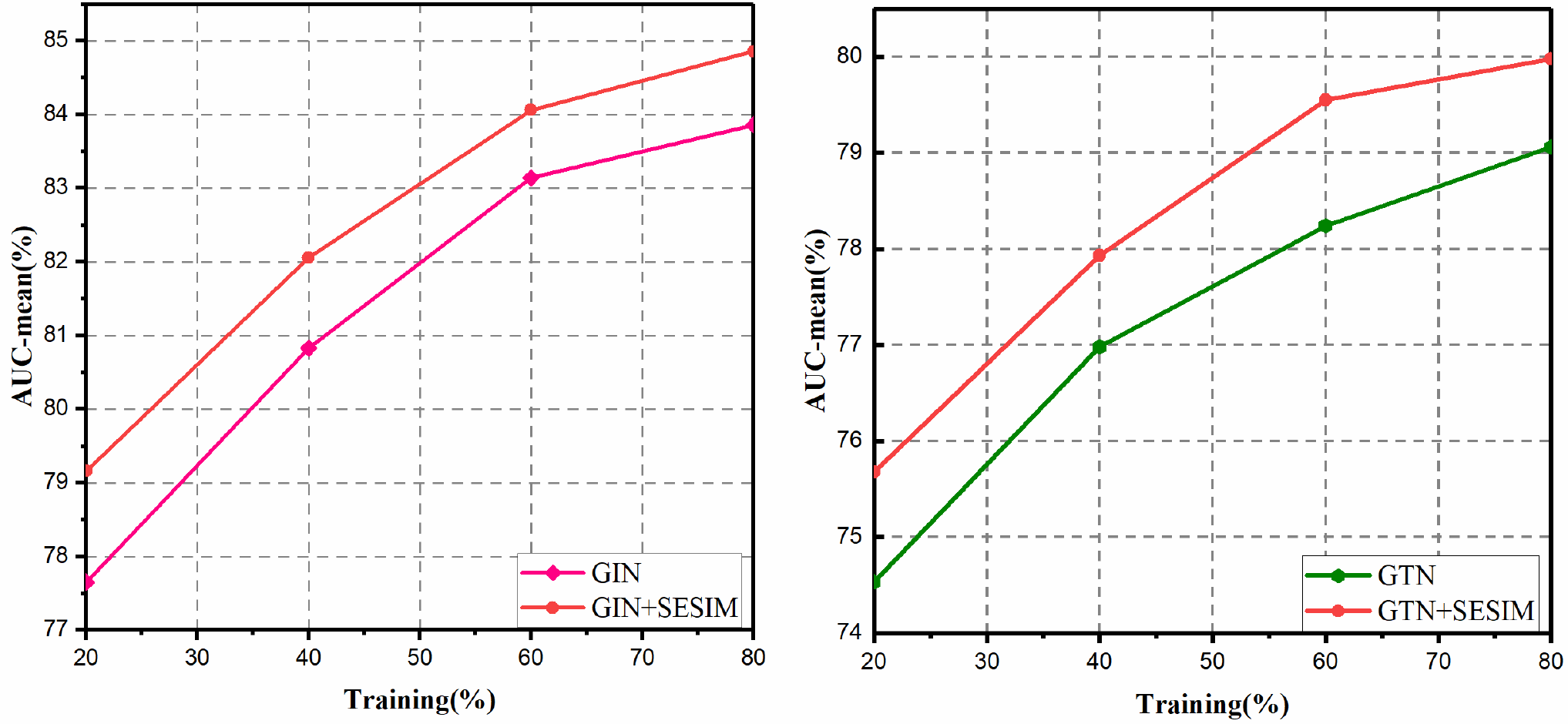}
    \end{minipage}
}
\subfigure[]
{
 	\begin{minipage}[b]{.8\linewidth}
        \centering
        \includegraphics[scale=0.67]{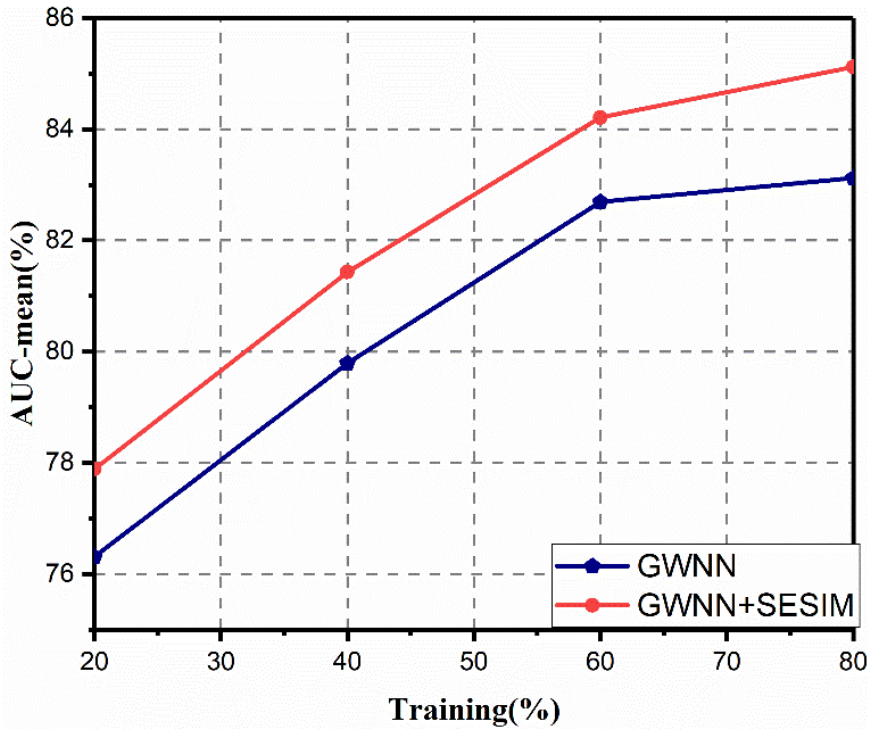}
    \end{minipage}
}
\caption{The AUC-mean results of different amount training data on Last-FM dataset}
\end{figure}

\subsection{ Node classification results and analysis}

\begin{table}[!htbp]
	\centering
	\label{tb7}
	\caption{The Macro-F1 node classification results}
	\begin{tabular}{cccccccc}
		\hline
		&\multicolumn{2}{c}{IMDB}  &\multicolumn{2}{c}{ACM} &\multicolumn{2}{c}{DBLP}  &\multirow{2}{*}{Average-increase} \\ \cline{1-7}
	 &Vanilla	&SESIM	&Vanilla	&SESIM	&Vanilla	&SESIM &	\\ \hline
GCN	&50.46	&52.67	&88.37	&91.68	&91.68	&92.17	&2.00 \\
GAT	&51.69	&52.98	&86.96	&87.86	&89.61	&92.98	&1.85 \\GIN	&49.65	&50.36	&85.37	&87.64	&88.62	&90.76	&1.71
\\GTN	&48.85	&51.68	&87.56	&90.94	&90.83	&93.25	&2.88
\\SGC	&50.97	&51.71	&86.79	&90.68	&88.68	&89.83	&1.93
\\FastGCN	&49.86	&50.84	&88.65	&89.74	&89.07	&91.87	&1.62 \\
GWNN	&50.76	&52.47	&87.06	&90.67	&90.14	&93.76	&2.98
\\Average- increase	&——	&1.49	&——	&2.64	&——	&2.36	 &—— \\ \hline    
	\end{tabular}
\end{table}

\begin{table}[!htbp]
	\centering
	\label{tb8}
	\caption{The Micro-F1 node classification results}
	\begin{tabular}{cccccccc}
		\hline
	&\multicolumn{2}{c}{IMDB}  &\multicolumn{2}{c}{ACM} &\multicolumn{2}{c}{DBLP}  &\multirow{2}{*}{Average-increase} \\ \cline{1-7}
	 &Vanilla	&SESIM	&Vanilla	&SESIM	&Vanilla	&SESIM &	\\ \hline
GCN	&53.76	&57.16	&88.07	&91.16	&92.59	&93.70	&2.53\\
GAT	&54.39	&55.81	&86.47	&87.73	&91.34	&93.17	&1.50
\\GIN	&52.86	&54.39	&84.76	&87.03	&90.38	&92.41	&1.94
\\GTN	&53.46	&55.71	&87.51	&90.74	&92.17	&93.97	&2.43
\\SGC	&55.37	&53.76	&86.27	&90.60	&90.05	&90.76	&1.14
\\FastGCN	&53.68	&54.97	&88.47	&89.38	&89.81	&92.83	&1.74 \\
GWNN	&52.71	&54.82	&86.91	&90.60	&91.86	&93.98	&2.64 \\
Average- increase	&——	&1.48	&——	&2.68	&——	&1.80	&—— \\ \hline 
	\end{tabular}
\end{table}

For the sake of testing the performance of our proposed model SESIM, we also verify it using a standard node classification task. We perform node classification task on the datasets from entertainment and academia fields, IMDB is a dataset of movie and television programs, ACM and DBLP are academic dataset. In this task, to evaluate the node classification results from different perspectives, Macro-F1 and Micro-F1 evaluation criteria are used. We also compute Average-increase of all baselines on the same dataset in the last line and Average-increase of each baseline on different datasets in the last column. 

All the experimental results are shown in Table 7 and Table 8. By observing these two tables, our model achieves state-of-the-art results on three datasets both in these evaluation criteria. Since the SESIM designed by predicting jump number on each metapath can steadily improve the ability of node classification for traditional neural networks models. The dominating reason for the improvements is pretext tasks can help models to better learn of graph structure, thus more effectively capturing the properties from their neighbor nodes for node classification task.

Comparing our proposed model SESIM applied to different traditional graph neural networks on different datasets, we also find difference in the performance of node classification task. Specifically, for these two   evaluation criteria, the results in Macro-F1 are better than in Micro-F1. From the perspective of dataset, SESIM achieves the best performance on ACM dataset among three datasets by 2.64\% and 2.68\% respectively of Average-increase in two evaluation criteria. We argue that this is due to the characteristics of the dataset itself, i.e., the supervised information on each metapath in more suitable for extracting the internal intrinsic connection on ACM dataset.

 In order to illustrate the performance of our model more intuitively, we plot the node classification accuracies in Macro-F1 in Fig.8, while the result in Micro-F1 no longer plotted. From the spacing between two points in each set of Fig.8(a), Fig.8(b) and Fig.8(c) and the Average-increase of Table 7 and Table 8, we can see the phenomenon that GCN and GWNN which belong to spectral-based GNNs after applied SESIM perform better with the Average-increase 2.00\% and 2.98\% respectively than some other spectral-based one. We guess that SESIM may greatly improve the performance of spectral-based GNNs for the ability of aggregating node features. We believe that this difference is due to the design of the pretext task, and in the future, we will explore other pretext tasks for achieving better effect.

\begin{figure}[htbp]
\centering

\subfigure[]
{
 	\begin{minipage}[b]{.8\linewidth}
        \centering
        \includegraphics[scale=0.96]{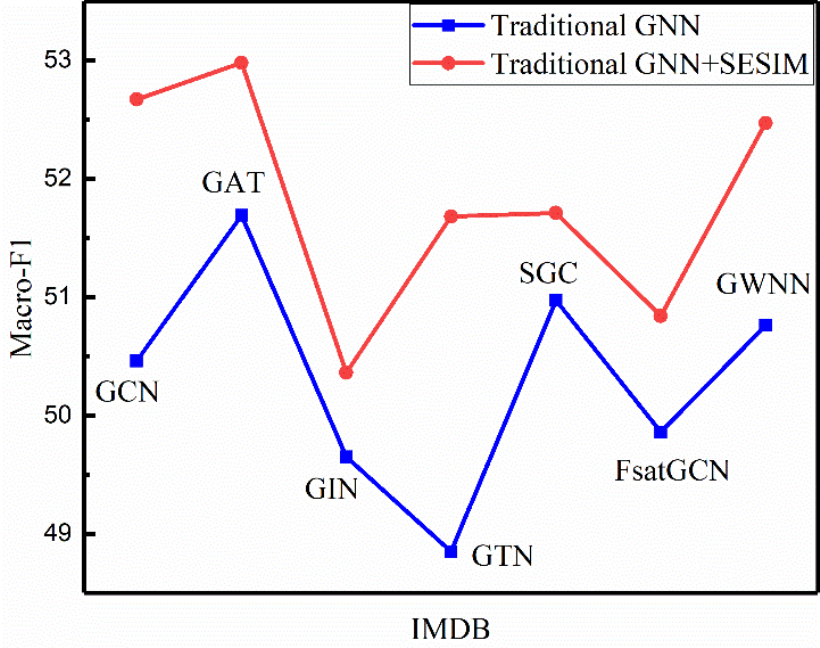}
    \end{minipage}
}
\subfigure[]
{
 	\begin{minipage}[b]{.8\linewidth}
        \centering
        \includegraphics[scale=0.96]{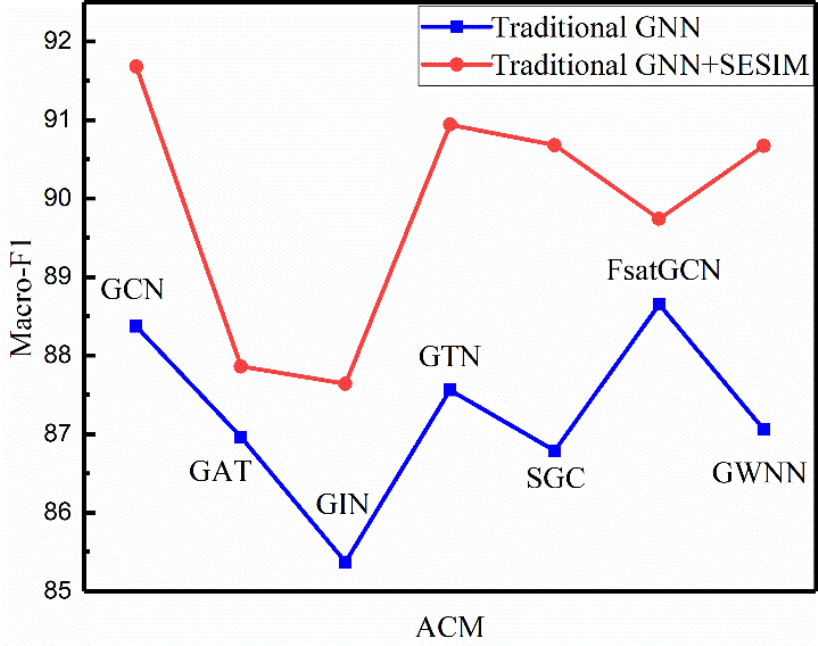}
    \end{minipage}
}
\subfigure[]
{
    \begin{minipage}[b]{.8\linewidth}
        \centering
        \includegraphics[scale=0.96]{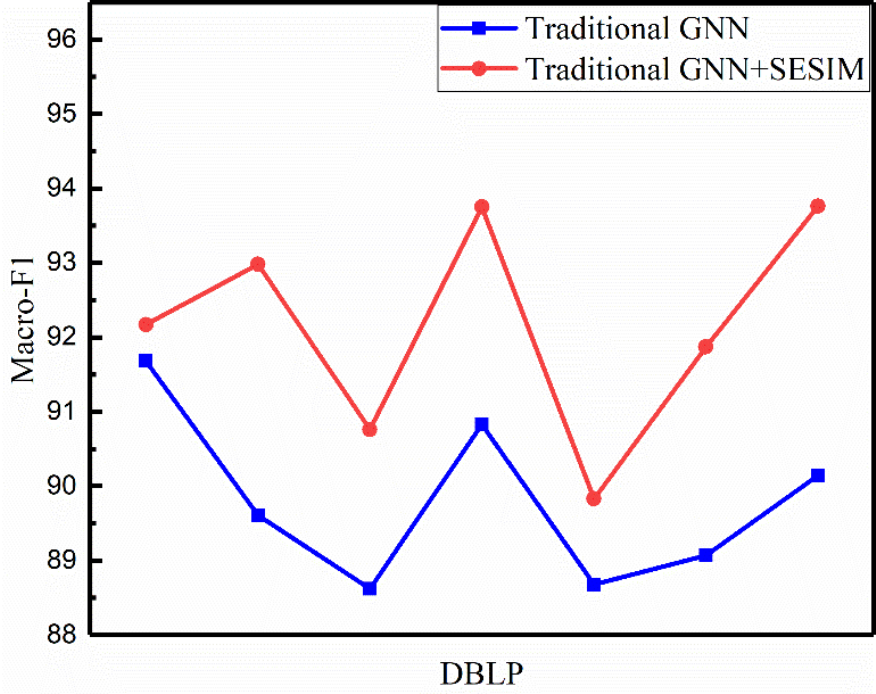}
    \end{minipage}
}
\caption{The Macro-F1 results}
\end{figure}

\subsection{Effect of metapath number}

In this section, one group of contrast experiments is implemented to evaluate the effect of the metapath number. We design 5 types of metapaths in subsection 5.5 and 5.6, e.g., user-item-user, item-user-item-user, user-item-artist.origin-item etc. Metapath is the important attribute in heterogeneous graph, we consider that metapath number have a huge impact on the pretext tasks during the training process of the model. Because in SESIM we design jump number prediction in each metapath, different numbers of metapaths would make the pretext tasks have different understandings of graph structure, thus making different impact on primary tasks.

We do contrast experiments using different numbers of metapaths by matrix operations of the adjacency matrix composed of each edge type. On Last-FM dataset, we design 3,4,5 and 6 types of metapaths. To verify the impact of different numbers of metapaths for experimental results, we compute the Average-increase for traditional graph neural networks on the Last-FM dataset after applied SESIM for link prediction task.  As depicted in Fig.9, compared with 3 types of metapaths, the Average-increase of 4 and 5 types of metapaths are increased by 0.16\% and 0.27\% respectively. The number of metapaths ranges from 3 to 5, and the overall trend of change is increasing, which shows metapath number have an impact on pretext tasks and more metapaths can abstract the relationship between nodes more clearly. While further differentiating 6 types of metapaths and higher types of metapaths would degrade the performance. We believe that there are overlapping parts between metapaths, which confuses the model.

\begin{figure}[!htbp]
	\centering
	\includegraphics[scale=1]{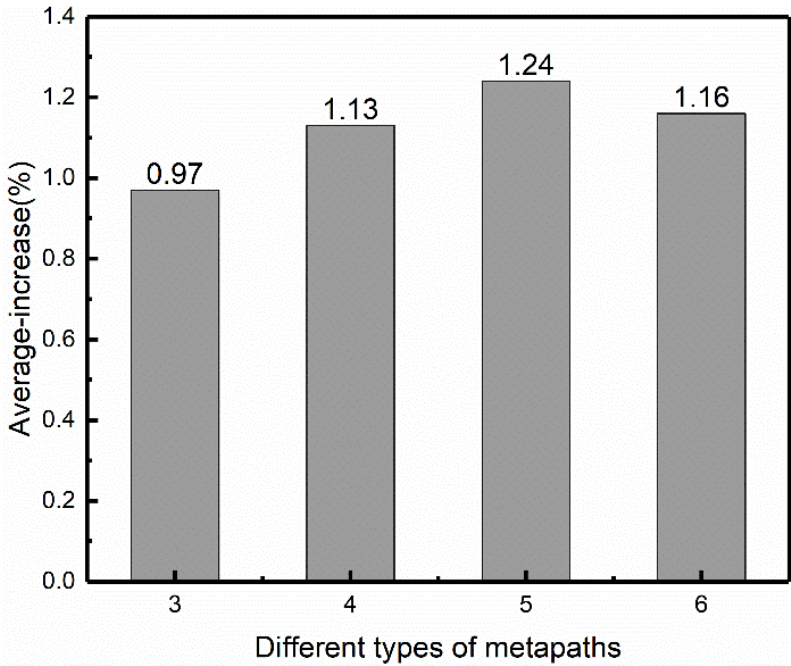}
	\caption{The Average-increase of different types of metapaths}
	\label{fig9}
\end{figure}

\subsection{Effect of different jump number}

In this subsection, one group of contrast experiments are conducted to evaluate the effect of the different jump number. The pretext tasks in our proposed model is jump number prediction in each metapath, jump number between nodes is natural structure information in heterogeneous graph.
 
In subsection 5.5 and 5.6, the max jump number we designed is 4. While the max jump numbers in our contrast experiments are set to 2,3,4 and 5 to explore the impact on downstream task. The experimental setup is the same as in subsection 5.7, we do those contrast experiments on Last-FM dataset and compute the Average-increase for traditional graph neural networks after applied SESIM for link prediction task. The experimental results are depicted in Fig.10. from Fig.10, we can get two observations. (1) 2,3 and 4 jump number are benefit for downstream task. As the jump number increases, the Average-increase accuracies of link prediction become higher and higher. (2) But when setting max jump number to 5 or more quantity, the index of experimental results drops. From these two phenomena, we are more convinced of the effectiveness of the pretext tasks we designed, because changes in the jump number can cause changes in the accuracies of the primary task. The explanation for these observations is that the max jump number of 2 or 3 for pretext tasks provide weak supervised information, which cannot effectively learn and use heterogeneous graph structure in each metapath. While too many jump numbers will make the model confusing and fuzzy, which pose adversely effect on the primary task. 

\begin{figure}[!htbp]
	\centering
	\includegraphics[scale=1]{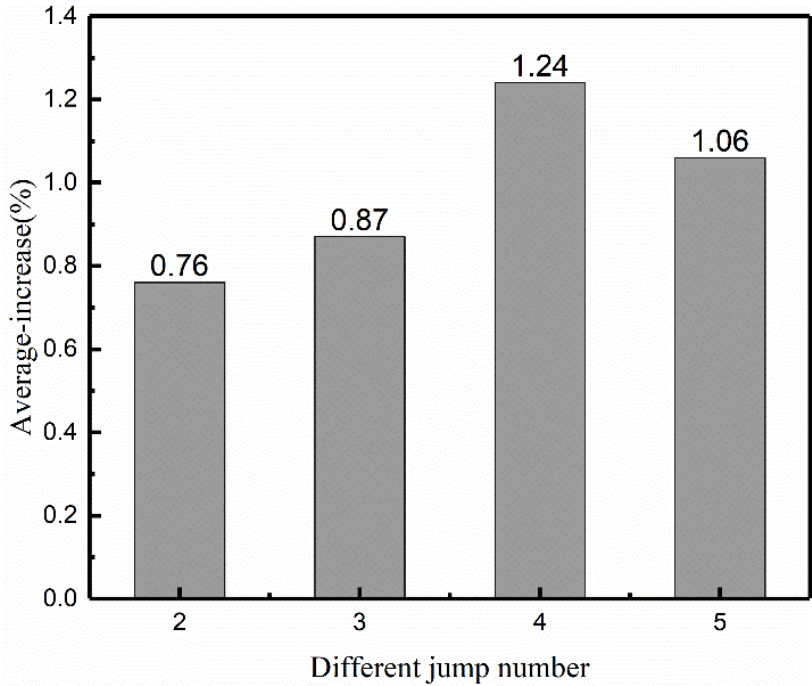}
	\caption{The Average-increase of different jump number}
	\label{fig10}
\end{figure}

\section{Conclusions and future work}

In the real-word, heterogeneous graph is ubiquitous, for the ability of which can model various types of nodes and diverse interactive information. Also, semi-supervised learning methods are used in studying heterogeneous graph commonly. But in some real-word environment, obtaining node labels is expensive and challenging. Recently, self-supervised learning becomes a promising vehicle for solving this problem, which can generate supervised information from the data itself spontaneously. So, the self-supervised learning on heterogeneous graphs is worth studying. In this paper, a self-supervised learning algorithm for heterogeneous graph via structure information based on metapath (SESIM) is proposed. SESIM is an effective self-supervised learning algorithm for heterogeneous graph, which uses graph structure to mine graph information by itself. SESIM constructs supervised information by predicting jump number between nodes in each metapath, which is used as pretext tasks to enhance the performance of primary task.  SESIM can not only solve the problem about data labels, but also mine structure information of heterogeneous graph, which enables graph to present rich information. Thus, SESIM can make traditional graph neural networks have better learning ability for aggregating nodes and utilizing graph structure. Experimental results verify that our proposed model outperforms the traditional approaches for link prediction and node classification task, and we discuss the effect of metapath number and jump number on prediction performance simultaneously. In the future, we will continue to study self-supervised learning on heterogeneous graph, exploring more valuable supervised information which comes from graph data itself.

\bibliography{mybibfile}

\begin{thebibliography}{10}
\expandafter\ifx\csname url\endcsname\relax
  \def\url#1{\texttt{#1}}\fi
\expandafter\ifx\csname urlprefix\endcsname\relax\def\urlprefix{URL }\fi
\expandafter\ifx\csname href\endcsname\relax
  \def\href#1#2{#2} \def\path#1{#1}\fi

\bibitem{1kim2014convolutional}
Y.~Kim, Convolutional neural networks for sentence classification, arXiv
  preprint arXiv:1408.5882 (2014).

\bibitem{2donahue2014decaf}
J.~Donahue, Y.~Jia, O.~Vinyals, J.~Hoffman, N.~Zhang, E.~Tzeng, T.~Darrell,
  Decaf: A deep convolutional activation feature for generic visual
  recognition, in: International conference on machine learning, PMLR, 2014,
  pp. 647--655.

\bibitem{3zhu2018learning}
H.~Zhu, X.~Li, P.~Zhang, G.~Li, J.~He, H.~Li, K.~Gai, Learning tree-based deep
  model for recommender systems, in: Proceedings of the 24th ACM SIGKDD
  International Conference on Knowledge Discovery \& Data Mining, 2018, pp.
  1079--1088.

\bibitem{4niepert2016learning}
M.~Niepert, M.~Ahmed, K.~Kutzkov, Learning convolutional neural networks for
  graphs, in: International conference on machine learning, PMLR, 2016, pp.
  2014--2023.

\bibitem{5gao2018large}
H.~Gao, Z.~Wang, S.~Ji, Large-scale learnable graph convolutional networks, in:
  Proceedings of the 24th ACM SIGKDD International Conference on Knowledge
  Discovery \& Data Mining, 2018, pp. 1416--1424.

\bibitem{6atwood2016diffusion}
J.~Atwood, D.~Towsley, Diffusion-convolutional neural networks, in: Advances in
  neural information processing systems, 2016, pp. 1993--2001.

\bibitem{7madjiheurem2019representation}
S.~Madjiheurem, L.~Toni, Representation learning on graphs: A reinforcement
  learning application (2019).

\bibitem{8lee2017transfer}
J.~Lee, H.~Kim, J.~Lee, S.~Yoon, Transfer learning for deep learning on
  graph-structured data, in: Proceedings of the AAAI Conference on Artificial
  Intelligence, Vol.~31, 2017.

\bibitem{9ying2019gnnexplainer}
R.~Ying, D.~Bourgeois, J.~You, M.~Zitnik, J.~Leskovec, Gnnexplainer: Generating
  explanations for graph neural networks (2019).

\bibitem{10hu2020strategies}
W.~Hu, B.~Liu, J.~Gomes, M.~Zitnik, P.~Liang, V.~Pande, J.~Leskovec, Strategies
  for pre-training graph neural networks, in: International Conference on
  Learning Representations (ICLR), 2020.

\bibitem{11davis2017comparative}
A.~P. Davis, C.~J. Grondin, R.~J. Johnson, D.~Sciaky, B.~L. King, R.~McMorran,
  J.~Wiegers, T.~C. Wiegers, C.~J. Mattingly, The comparative toxicogenomics
  database: update 2017, Nucleic acids research 45~(D1) (2017) D972--D978.

\bibitem{12wang2019heterogeneous}
X.~Wang, H.~Ji, C.~Shi, B.~Wang, Y.~Ye, P.~Cui, P.~S. Yu, Heterogeneous graph
  attention network, in: The World Wide Web Conference, 2019, pp. 2022--2032.

\bibitem{13fu2020magnn}
X.~Fu, J.~Zhang, Z.~Meng, I.~King, Magnn: Metapath aggregated graph neural
  network for heterogeneous graph embedding, in: Proceedings of The Web
  Conference 2020, 2020, pp. 2331--2341.

\bibitem{14hu2020heterogeneous}
Z.~Hu, Y.~Dong, K.~Wang, Y.~Sun, Heterogeneous graph transformer, in:
  Proceedings of The Web Conference 2020, 2020, pp. 2704--2710.

\bibitem{15pathak2016context}
D.~Pathak, P.~Krahenbuhl, J.~Donahue, T.~Darrell, A.~A. Efros, Context
  encoders: Feature learning by inpainting, in: Proceedings of the IEEE
  conference on computer vision and pattern recognition, 2016, pp. 2536--2544.

\bibitem{16vondrick2018tracking}
C.~Vondrick, A.~Shrivastava, A.~Fathi, S.~Guadarrama, K.~Murphy, Tracking
  emerges by colorizing videos, in: Proceedings of the European conference on
  computer vision (ECCV), 2018, pp. 391--408.

\bibitem{17peng2020self}
Z.~Peng, Y.~Dong, M.~Luo, X.-M. Wu, Q.~Zheng, Self-supervised graph
  representation learning via global context prediction, arXiv preprint
  arXiv:2003.01604 (2020).

\bibitem{18rong2020self}
Y.~Rong, Y.~Bian, T.~Xu, W.~Xie, Y.~Wei, W.~Huang, J.~Huang, Self-supervised
  graph transformer on large-scale molecular data, arXiv preprint
  arXiv:2007.02835 (2020).

\bibitem{19hwang2020self}
D.~Hwang, J.~Park, S.~Kwon, K.-M. Kim, J.-W. Ha, H.~J. Kim, Self-supervised
  auxiliary learning with meta-paths for heterogeneous graphs, arXiv preprint
  arXiv:2007.08294 (2020).

\bibitem{20wang2021self}
P.~Wang, K.~Agarwal, C.~Ham, S.~Choudhury, C.~K. Reddy, Self-supervised
  learning of contextual embeddings for link prediction in heterogeneous
  networks, in: Proceedings of the Web Conference 2021, 2021.

\bibitem{21bruna2014spectral}
J.~Bruna, W.~Zaremba, A.~Szlam, Y.~LeCun, Spectral networks and locally
  connected networks on graphs (2014).

\bibitem{22defferrard2016convolutional}
M.~Defferrard, X.~Bresson, P.~Vandergheynst, Convolutional neural networks on
  graphs with fast localized spectral filtering, in: NIPS, 2016.

\bibitem{23kipf2017semi}
T.~N. Kipf, M.~Welling, Semi-supervised classification with graph convolutional
  networks (2017).

\bibitem{24xu2019graph}
B.~Xu, H.~Shen, Q.~Cao, Y.~Qiu, X.~Cheng, Graph wavelet neural network, arXiv
  preprint arXiv:1904.07785 (2019).

\bibitem{25hamilton2017inductive}
W.~L. Hamilton, R.~Ying, J.~Leskovec, Inductive representation learning on
  large graphs, in: Proceedings of the 31st International Conference on Neural
  Information Processing Systems, 2017, pp. 1024--1034.

\bibitem{26velivckovic2018graph}
P.~Veli{\v{c}}kovi{\'c}, A.~Casanova, P.~Lio, G.~Cucurull, A.~Romero,
  Y.~Bengio, Graph attention networks (2018).

\bibitem{27zhao2020network}
J.~Zhao, X.~Wang, C.~Shi, Z.~Liu, Y.~Ye, Network schema preserved heterogeneous
  information network embedding, in: 29th International Joint Conference on
  Artificial Intelligence (IJCAI), 2020.

\bibitem{28doersch2015unsupervised}
C.~Doersch, A.~Gupta, A.~A. Efros, Unsupervised visual representation learning
  by context prediction, in: Proceedings of the IEEE international conference
  on computer vision, 2015, pp. 1422--1430.

\bibitem{29pathak2016context}
D.~Pathak, P.~Krahenbuhl, J.~Donahue, T.~Darrell, A.~A. Efros, Context
  encoders: Feature learning by inpainting (2016).

\bibitem{30sermanet2018time}
P.~Sermanet, C.~Lynch, Y.~Chebotar, J.~Hsu, E.~Jang, S.~Schaal, S.~Levine,
  G.~Brain, Time-contrastive networks: Self-supervised learning from video, in:
  2018 IEEE international conference on robotics and automation (ICRA), IEEE,
  2018, pp. 1134--1141.

\bibitem{31wang2015unsupervised}
X.~Wang, A.~Gupta, Unsupervised learning of visual representations using
  videos, in: Proceedings of the IEEE international conference on computer
  vision, 2015, pp. 2794--2802.

\bibitem{32hjelm2018learning}
R.~D. Hjelm, A.~Fedorov, S.~Lavoie-Marchildon, K.~Grewal, P.~Bachman,
  A.~Trischler, Y.~Bengio, Learning deep representations by mutual information
  estimation and maximization, arXiv preprint arXiv:1808.06670 (2018).

\bibitem{33he2020momentum}
K.~He, H.~Fan, Y.~Wu, S.~Xie, R.~Girshick, Momentum contrast for unsupervised
  visual representation learning, in: Proceedings of the IEEE/CVF Conference on
  Computer Vision and Pattern Recognition, 2020, pp. 9729--9738.

\bibitem{34jin2020self}
W.~Jin, T.~Derr, H.~Liu, Y.~Wang, S.~Wang, Z.~Liu, J.~Tang, Self-supervised
  learning on graphs: Deep insights and new direction, arXiv preprint
  arXiv:2006.10141 (2020).

\bibitem{35shu2019meta}
J.~Shu, Q.~Xie, L.~Yi, Q.~Zhao, S.~Zhou, Z.~Xu, D.~Meng, Meta-weight-net:
  Learning an explicit mapping for sample weighting, arXiv preprint
  arXiv:1902.07379 (2019).

\bibitem{36wang2019knowledge}
H.~Wang, F.~Zhang, M.~Zhang, J.~Leskovec, M.~Zhao, W.~Li, Z.~Wang,
  Knowledge-aware graph neural networks with label smoothness regularization
  for recommender systems, in: Proceedings of the 25th ACM SIGKDD international
  conference on knowledge discovery \& data mining, 2019, pp. 968--977.

\bibitem{37wang2018ripplenet}
H.~Wang, F.~Zhang, J.~Wang, M.~Zhao, W.~Li, X.~Xie, M.~Guo, Ripplenet:
  Propagating user preferences on the knowledge graph for recommender systems,
  in: Proceedings of the 27th ACM International Conference on Information and
  Knowledge Management, 2018, pp. 417--426.

\bibitem{38xu2018powerful}
K.~Xu, W.~Hu, J.~Leskovec, S.~Jegelka, How powerful are graph neural networks?,
  arXiv preprint arXiv:1810.00826 (2018).

\bibitem{39yun2019graph}
S.~Yun, M.~Jeong, R.~Kim, J.~Kang, H.~J. Kim, Graph transformer networks,
  Advances in Neural Information Processing Systems 32 (2019) 11983--11993.

\bibitem{40wu2019simplifying}
F.~Wu, A.~Souza, T.~Zhang, C.~Fifty, T.~Yu, K.~Weinberger, Simplifying graph
  convolutional networks, in: International conference on machine learning,
  PMLR, 2019, pp. 6861--6871.

\bibitem{41chen2018fastgcn}
J.~Chen, T.~Ma, C.~Xiao, Fastgcn: fast learning with graph convolutional
  networks via importance sampling, arXiv preprint arXiv:1801.10247 (2018).

\bibitem{42kingma2015adam}
D.~Kingma, L.~Ba, et~al., Adam: A method for stochastic optimization (2015).

\end{thebibliography}

\end{document}